# Comparative Assessment of Deep Learning Architectures for Underwater Subsurface Kelp Forest Segmentation with The Kelp-o-Tron

Sundarabalan Balasubramanian[1*], César Borja[2], Ana C. Murillo[2], Lexi N. Wilkes[1], Meredith L. McPherson[1], Kira A. Krumhansl[3], Jennifer A. Dijkstra[4], Jarrett E. K. Byrnes[1]

1 - Department of Biology, University of Massachusetts Boston, Boston, MA 02125 USA
2 - DIIS-i3A, University of Zaragoza, Zaragoza, Spain
3 - Bedford Institute of Oceanography, Fisheries and Oceans Canada, Dartmouth, NS, B3A 2T2 Canada
4 - Center for Coastal and Ocean Mapping, University of New Hampshire, Durham, NH 03824, USA

*Corresponding author email: Velaudhaperumalbalas@umb.edu

**Abstract:**

Submerged kelp forests are vital coastal ecosystems that support marine biodiversity and ecosystem dynamics, yet accurate underwater kelp segmentation remains challenging due to optical degradation, illumination variability, turbidity, overlapping vegetation, and complex benthic backgrounds. We systematically evaluated three deep learning semantic segmentation frameworks, ResNet34-U-Net, ResNet50-DeepLabV3, and a hybrid ResNet50-ASPP-Transformer architecture, for kelp detection using high-resolution underwater RGB imagery collected from northeastern U.S. coastal waters. A dataset of 3,395 SSeg assisted annotated image-mask pairs was developed for model training and validation, while geographically independent sites were used for quantitative and qualitative evaluation. All models used consistent preprocessing, augmentation, and evaluation protocols. On independent test data, ResNet50-DeepLabV3 achieved the highest Dice (0.7120) and Intersection over Union (IoU; 0.6267), followed by ResNet34 U Net (Dice 0.6868; IoU 0.5978). The hybrid ASPP Transformer achieved the highest pixel accuracy (0.8528) but lower Dice (0.6437) and IoU (0.5746). External qualitative evaluation further showed that DeepLabV3 produced more consistent segmentation across varying environmental conditions, image qualities, and benthic habitats. Overall, ResNet50-DeepLabV3, termed Kelp-O-Tron, provided the best balance of segmentation accuracy, robustness, and generalization. The dataset, annotation workflow, and comparative evaluation provide resources for advancing automated underwater habitat mapping and ecological monitoring.

## 1. Introduction

Kelp forests are among the most productive marine ecosystems (Mann, 1973; Byrnes et al., 2011; Krause-Jensen and Duarte, 2016). These forests of large brown macroalgae from the family Laminariales distributed throughout temperate coastal regions worldwide covering nearly a quarter of the world's coastlines (Steneck et al., 2002, Krumhansl et al., 2016). They support diverse marine communities and are of particular importance for coastal biodiversity (Bartsch et al., 2008; Byrnes et al., 2011; Dayton, 1985; Graham et al., 2007; Hamilton et al., 2020). Kelp forests provide significant ecosystem services such as primary production, nutrient cycling, shoreline protection, and carbon sequestration, and thereby contributing to blue-carbon storage and climate regulation (Eger et al., 2022; McHenry et al., 2025). However, kelp ecosystems face increasing threats from environmental and anthropogenic stressors, including ocean warming, marine heatwaves, pollution, overgrazing, and coastal development, resulting in substantial declines in kelp distribution and abundance across numerous coastal regions (Bell et al., 2020; Sato et al., 2025). Assessment of the distribution and dynamics of kelps is necessary to improve the understanding of ecosystem change and support marine conservation and management strategies (Eger et al., 2025; Filbee-Dexter et al., 2020). While remote sensing has provided a robust solution for surface canopy forming kelps (Bell et al., 2023; 2020; Cavanaugh et al., 2010: Moro soto et al., 2020; Schroeder et al., 2020), the majority of the world's kelps do not reach the surface and consequently mapping efforts at large scales rely on underwater imagery.

Underwater imagery has become an increasingly important tool for marine ecological monitoring, enabling the assessment of habitat composition, species distributions, and ecosystem condition across broad spatial and temporal scales. However, the interpretation of underwater imagery remains challenging because image quality is often degraded by light attenuation, scattering, turbidity, and water-column variability, which reduce image clarity, contrast, and color fidelity (Schechner and Karpel, 2005; Akkaynak and Treibitz, 2019). In addition, large-scale monitoring programs using underwater cameras, remotely operated vehicles (ROVs), and autonomous platforms generate vast quantities of imagery that require extensive processing and interpretation. Traditional annotation approaches rely heavily on expert review and manual delineation of habitat features, making the extraction of ecological information labor-intensive, time-consuming, and difficult to scale across large datasets (Durden et al., 2016; González-Rivero et al., 2020). Consequently, automated image-analysis techniques have become increasingly important for efficiently converting underwater imagery into quantitative ecological information for habitat monitoring, biodiversity assessment, and conservation planning.

To address these challenges, computer vision and deep learning have become powerful tools for automated analysis of ecological imagery, enabling efficient detection, classification, and quantification of organisms and habitats (Weinstein, 2018; Borowiec et al., 2022; Game et al., 2026). These advances can help address biodiversity knowledge gaps and have increasingly supported automated classification and mapping of marine habitats from underwater imagery (Pollock et al., 2025). Mahmood et al. (2020) showed that deep image representations substantially outperformed traditional handcrafted image features for automated classification of the kelp species *Ecklonia radiata* from underwater imagery at Rottnest Island, Western Australia, highlighting the potential of deep learning for large-scale marine ecological monitoring. Williams et al. (2019) applied CoralNet to automate coral reef image annotation across American Samoa and the Main Hawaiian Islands, demonstrating that automated coral cover estimates closely matched human annotations and could support large-scale reef monitoring. González-Rivero et al. (2020) further demonstrated the use of deep learning and large-scale image analysis to quantify coral reef benthic composition across global monitoring programs, achieving expert-level accuracy while enabling rapid, cost-effective, and standardized reef monitoring. Pierce et al. (2020, 2021) extended automated coral reef analysis using semantic segmentation, demonstrating its potential to reduce annotation effort and support classification of three-dimensional reef models derived from structure-from-motion imagery. More recently, Noman et al. (2024) applied deep learning-based object detection models to identify the seagrass species *Halophila ovalis* from underwater imagery in the ECUHO datasets, demonstrating accurate and efficient automated detection of submerged aquatic vegetation for large-scale habitat monitoring. Collectively, these studies highlight the growing role of computer vision in marine ecology, enabling scalable and repeatable analysis of underwater imagery across diverse coastal ecosystems.

Despite these advances, accurate segmentation of kelp forests remains considerably more challenging than many conventional marine habitat mapping applications. Kelp canopies exhibit highly dynamic morphology, with flexible fronds that change shape and orientation in response to currents, waves, and viewing geometry (Graham et al., 2007; Wernberg et al., 2019). Underwater imaging conditions further complicate segmentation through variable illumination, turbidity, backscatter, and wavelength-dependent light attenuation (Schechner and Karpel, 2005; Jaffe, 2015; Akkaynak and Treibitz, 2019). In addition, kelp forests often contain dense assemblages of understory and subcanopy macroalgae that can exhibit visual characteristics similar to canopy-forming kelp species, making species discrimination particularly challenging in complex three-dimensional habitats (Graham et al., 2007; Teagle et al., 2017; Wernberg et al., 2019). Consequently, segmentation approaches developed for general underwater scene analysis may not directly generalize to kelp-dominated environments. These challenges are further exacerbated by frequent occlusion, partial visibility of fronds, and large variations in kelp density and growth form across sites and seasons. These limitations highlight the need for robust and adaptive deep learning frameworks capable of handling underwater optical degradation, complex kelp morphology, and multi-scale spatial variability for accurate kelp segmentation.

Here, we evaluate the performance of various deep learning-based semantic segmentation frameworks for underwater kelp detection under challenging environmental conditions. Using a supervised learning approach, the networks were trained on underwater RGB images paired with manually generated or SSeg-assisted ground-truth segmentation masks. We perform a comparative assessment of three segmentation architectures representing different contextual learning strategies: ResNet34-U-Net, ResNet50-DeepLabV3, and a hybrid ASPP-Transformer framework. These architectures integrate encoder–decoder convolutional networks, multi-scale contextual feature extraction, and transformer-based global feature learning for semantic segmentation of underwater kelp imagery. Further we demonstrate how to use SSeg-assisted workflows to create ground-truth data for modeling with high-resolution underwater RGB images annotated initially with point counts rather than segmentation.

## 2. Datasets

Broadly, we collected underwater imagery to both develop the model and then used pre-existing data to test the model's out of sample performance. All imagery data was collected in New England, USA and Nova Scotia, Canada. Data consisted of images of rocky reefs between 1m and 20m in depth. The dataset includes substantial variability in kelp density, illumination conditions, turbidity, benthic community composition, and underwater scene complexity, providing a robust benchmark for evaluating segmentation model generalization across diverse underwater environments. In all data sets we focused on annotating and generating segmentations of the kelps *Saccharina latissima* and the *Laminaria digitata/Hedophyllum nigripes* complex. While *Agarum clathratum* was present in some images, it was negligible in cover. *Alaria esculenta* was not present. Images contained a diverse suite of other algae, and some images contained seagrass. The diversity of other similar species ensured that our classifier was developed to differentiate kelp from other benthic species.

### 2.1 Development dataset

To build an imagery dataset for model development, we extracted RGB images from video surveys acquired using towed camera systems across temperate reef ecosystems in New England, USA (Figure 1, Supplementary Table 1). Video data were collected with a GoPro HERO7 during 2022 surveys and a SpotX SquidPro during 2024 surveys. All imagery was captured at 1920 × 1080 pixels. GPS coordinates were recorded simultaneously during image acquisition using a Garmin GPS MAP 65 to document the geographic location of each underwater observation.

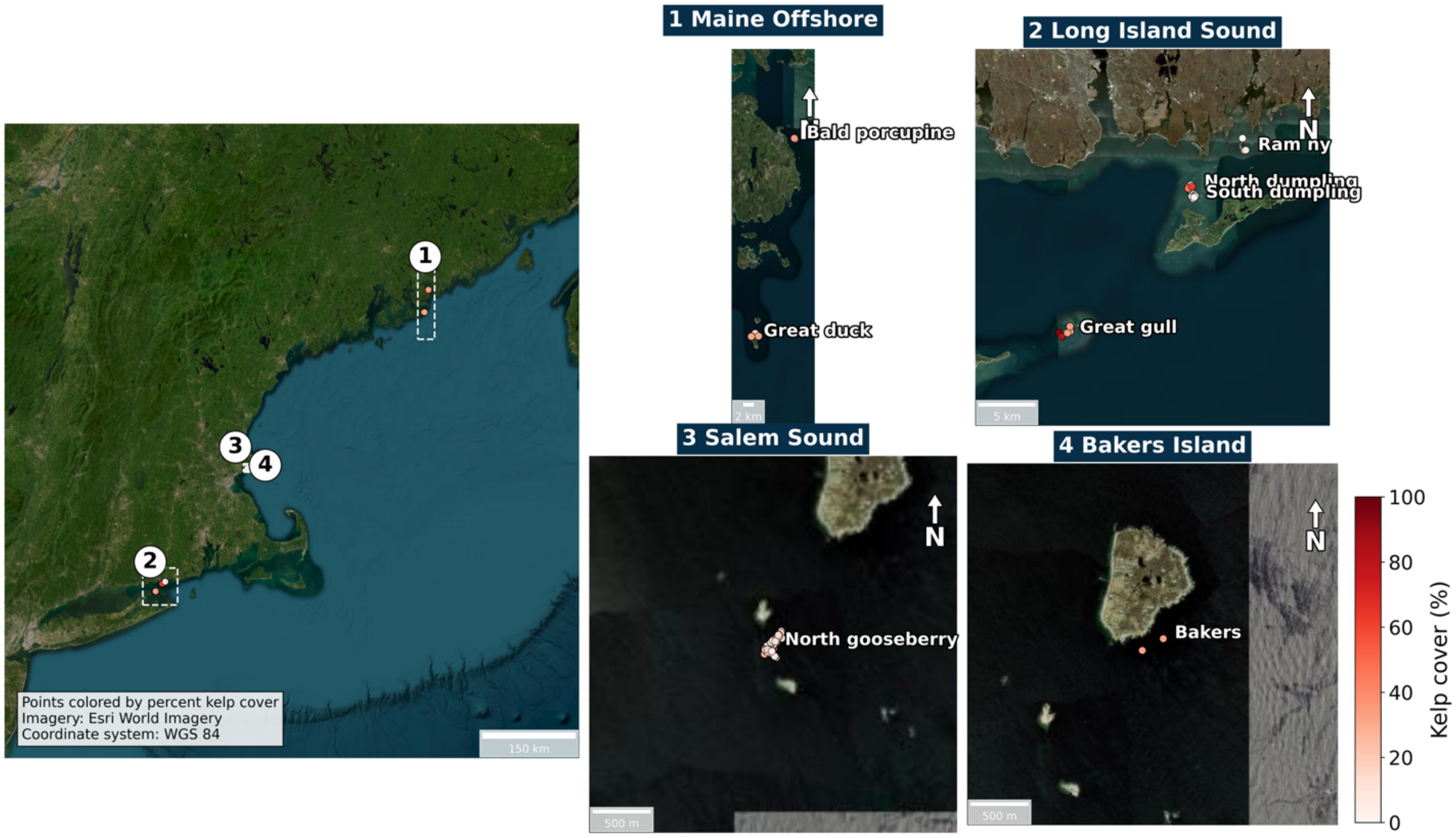

**Figure 1**. Spatial coverage of the Kelp underwater imagery dataset across New England, USA. The dataset includes georeferenced image collections from four major regions: Maine Offshore, Long Island Sound, Salem Sound, and Bakers Island. Sampling points are colored according to mean kelp percent cover (%) calculated from image-based annotations and used for model development and ecological analyses.

### 2.2. Evaluation dataset

We tested the fit model on three separate subtidal imagery datasets. The first was from a subset of imagery taken from the same sites as the development data during the summer of 2022 (shown in Figure 1, Supplementary Table 1). Fifty images from each site were selected to represent diverse underwater environmental conditions and varying kelp cover characteristics for robust model evaluation.

### 2.3. External dataset

To evaluate model transferability beyond the environments represented in the development dataset, two independent external underwater imagery datasets were used. First, we used underwater kelp imagery collected by the Canadian Department of Fisheries and Oceans (DFO) from Charlotte Island (45.313 °N, -60.968 °W, N = 43 images) and Woody Island (44.4498 °N, -63.7172°W, N = 37 images) off the coast of Nova Scotia, Canada (Supplementary Table 2). This data set was collected using an underwater drop camera system consisting of a downward facing GoPro HERO10 fixed to a metal frame. The camera system was towed along two transects parallel to shore at each site, one between 2-5 m depth and the second between 7-12 m depth. The camera was positioned between 1-3 m above the bottom, depending on the water depth and visibility. Still images were taken at 5 second intervals along each transect. Additional details of the survey design and image-collection methodology have been previously documented (Krumhansl et al., 2024). This data set also contained the invasive *Fucus serratus*, which provided an extra challenge for the model as there were no similar species in the training data, testing generality.

Second, model transferability was further assessed using an independent underwater imagery dataset collected from coastal New Hampshire, USA (N = 50 images). Underwater photographs of twenty 0.25 m² quadrats were collected at each of four study sites around the Isles of Shoals (Babb's Cove, Haley's Cove, Lunging Island, and White Island Cove) in 2014 using a GoPro Hero3 Black camera (Dijkstra et al. 2017). Quadrat images were obtained at depths ranging from 2.5 to 8 m, with the majority collected at approximately 8 m depth. The kelp assemblage at the study sites was dominated by *Saccharina latissima* and *Agarum clathratum*, with some *Laminaria digitata* and *Saccorhiza dermatodea*.

**3. Annotation and Image Pre-Processing**

For development data, we employed two separate annotation strategies. 623 images from the Gooseberry Islands, Massachusetts in 2022 had all kelps segmented using v7Darwin (https://www.v7darwin.com/). All other images were annotated using Semantic Segmentation (SSeg) method (Borja et al., 2026), an interactive annotation approach designed to transform sparse point annotations into dense segmentation masks.

SSeg combines an active sampling strategy that leverages both Segment Anything Model 2 (SAM2) and Superpixel based segmentation methods (Raine et al., 2022). SSeg facilitates semi-automatic object segmentation by propagating points or bounding boxes, placed by the user, into dense segmentation masks. This process reduces the need for manual pixel-wise annotation of underwater kelp imagery (Alonso et al., 2021; Borja et al., 2026). Within this workflow, SSeg enables an adapted application of SAM2 to underwater kelp imagery, while the ability to segment previously unseen images without task-specific model retraining is derived from the zero-shot capabilities of the underlying SAM2 foundation model.

Initially, underwater RGB images collected from kelp survey sites were imported into the annotation workflow. Positive foreground points and bounding boxes were manually provided over visible kelp regions, while negative prompts were occasionally applied over non-kelp structures such as rocks, sand, water-column artifacts, shadows, and benthic background regions to improve foreground-background separation (Hong et al., 2024). Based on these prompts, SSeg generated multiple candidate segmentation masks outlining the kelp structures within each image.

The generated masks were visually inspected and manually refined where necessary to improve boundary accuracy and remove false detections caused by complex benthic backgrounds. Additional interactive prompts were iteratively applied in challenging regions to correct under-segmentation and over-segmentation errors. This semi-automatic annotation strategy substantially accelerated the preparation of training datasets compared to fully manual polygon-based annotation approaches while maintaining high-quality segmentation masks suitable for supervised deep learning model training. The finalized masks were exported as binary segmentation images, where kelp pixels were assigned a foreground label (1) and all remaining pixels were assigned as background (0). The kelp cover percentage for each image was calculated as the ratio between the number of kelp pixels and the total number of pixels within the image, multiplied by 100. This metric represents the fractional image area occupied by kelp and was used to characterize kelp abundance variability across the dataset. Representative RGB underwater imagery and the corresponding SSeg-assisted segmentation masks are shown in Fig. 2.

Each underwater image was paired with a corresponding binary segmentation mask representing kelp and background regions for supervised semantic segmentation tasks. In total, 3395 underwater image-mask pairs with varying kelp coverage conditions were used for model development. Validation and test sites were selected independently to evaluate cross-site model generalization and to minimize spatial data leakage between dataset subsets.

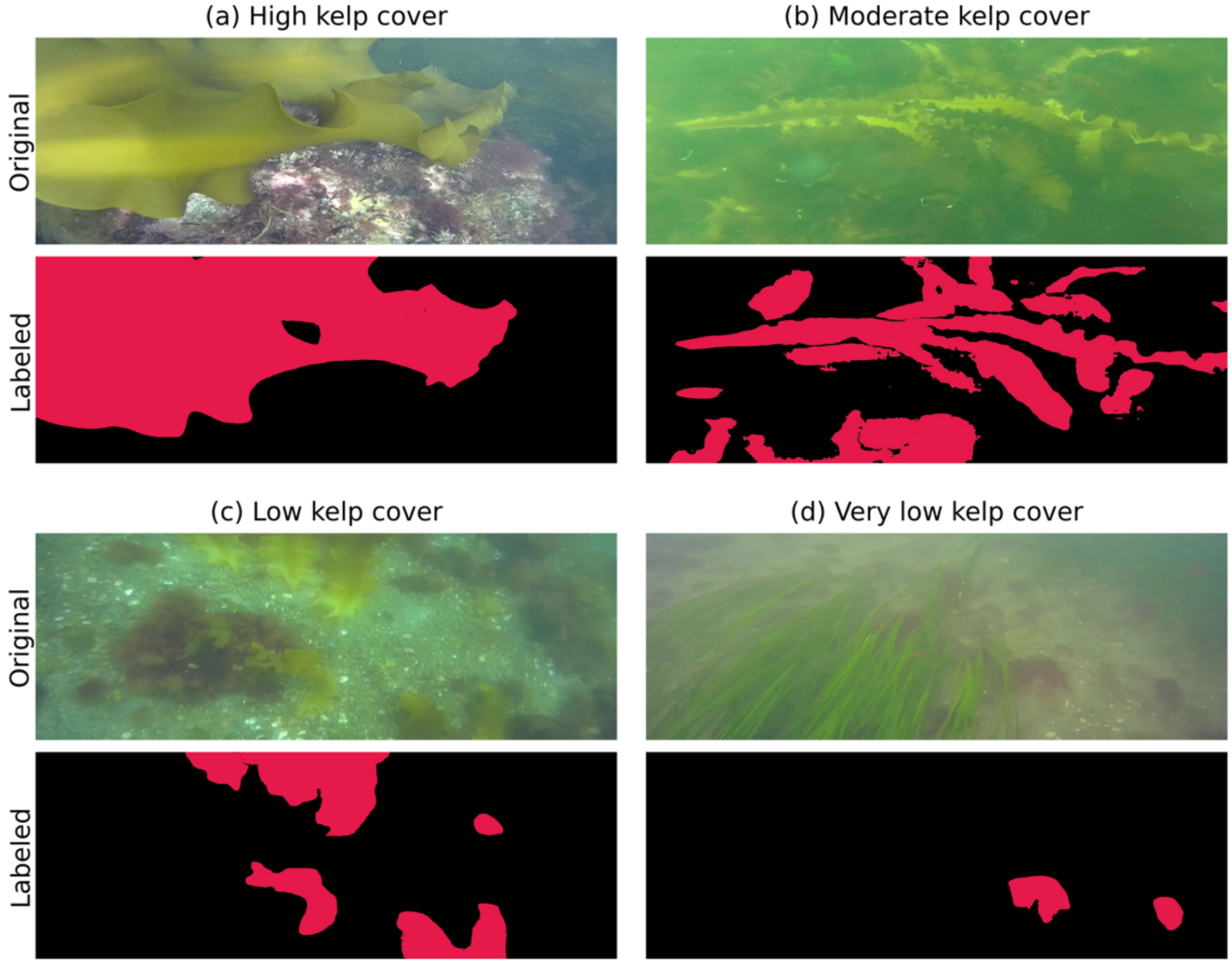

**Figure 2.** Representative underwater kelp images and corresponding ground-truth segmentation masks illustrating different levels of kelp cover: (a) high, (b) moderate, (c) low, and (d) very low. For each example, the upper panel shows the original RGB image, and the lower panel shows the corresponding SSeg-assisted annotated ground-truth mask, with kelp regions highlighted in pink. Note that in the very low kelp cover, the image is dominated by seagrass, which is not labeled.

## 4. Methodology

To achieve the objective of evaluating deep learning models for underwater kelp segmentation, a structured methodology was developed that includes dataset preparation, annotation, patch-based sampling, model training, and quantitative evaluation, as illustrated in Fig. 3. The dataset was divided into a site-wise train, validation, and test split to maintain independent evaluation across different locations. Image patches of size 512 × 512 pixels were extracted for model training. Three semantic segmentation models, namely ResNet34-U-Net, ResNet50-DeepLabV3, and ASPP-Transformer, were trained to generate binary kelp segmentation masks. Model performance was evaluated using Pixel Accuracy, Intersection over Union (IoU), Precision, Recall, F1-score, and Dice coefficient for comparative assessment.

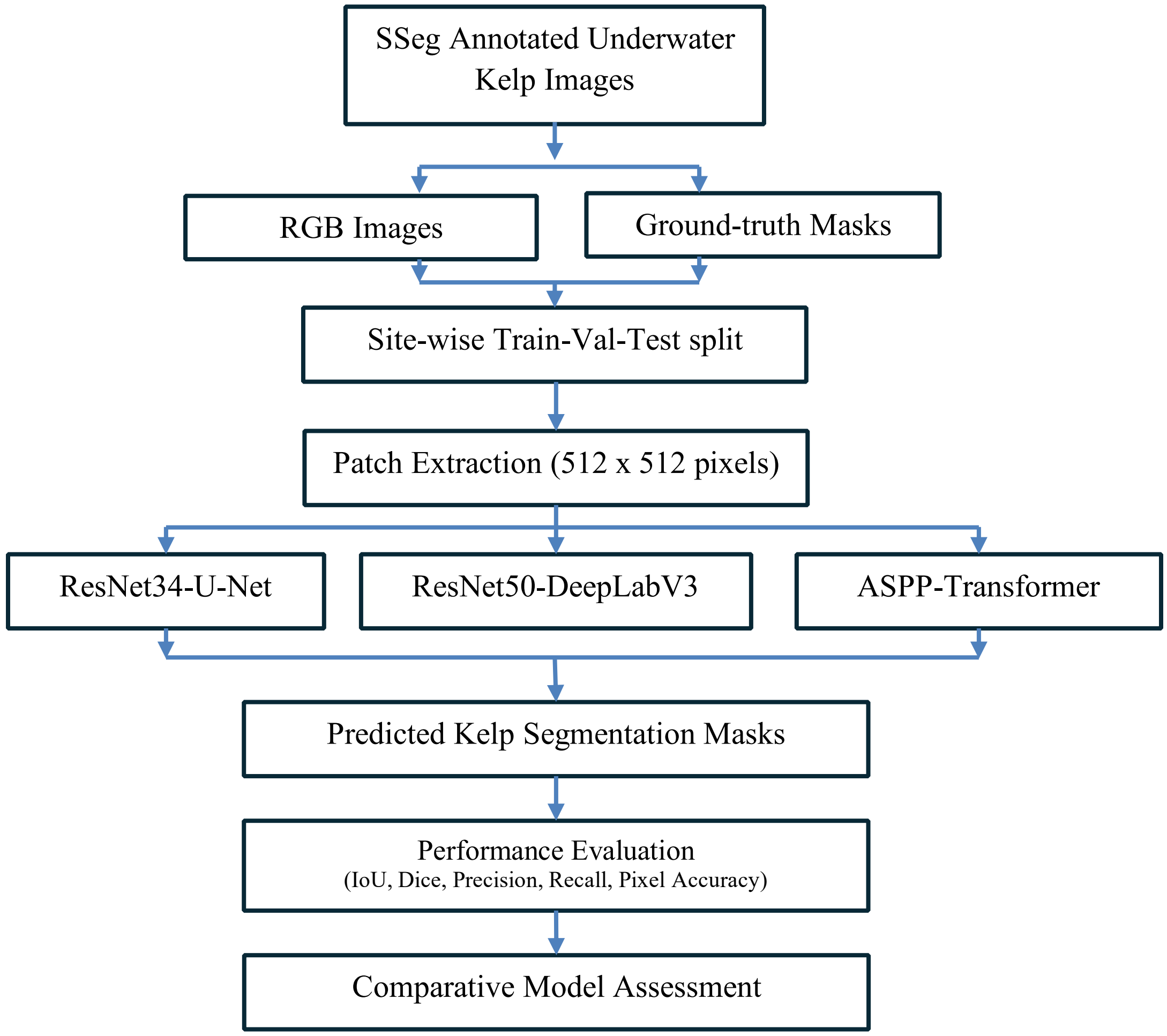

**Figure 3.** Workflow of the proposed framework for underwater kelp segmentation and comparative evaluation of three deep learning architectures: ResNet34-U-Net, ResNet50-DeepLabV3, and ASPP-Transformer.

### 4.1. Data Preprocessing

The underwater kelp dataset contains several challenges caused by differences in environmental conditions and image acquisition geometry that must be addressed prior to model training (Anwar and Li, 2020; Li et al., 2018; Torralba and Efros, 2011). To ensure robust model generalization across different geographic locations, a structured preprocessing workflow was designed, consisting of site-wise dataset splitting, patch-based sampling, and class-balanced patch selection.

#### *4.1.1. Site-wise split strategy*

The first stage of preprocessing involved a site-wise dataset split designed to ensure robust model generalization across different geographic locations. Instead of applying a random image-wise split, the dataset was divided according to acquisition sites to avoid data leakage and to enable a more reliable evaluation of the model's ability to generalize to unseen environments, as recommended in recent deep learning studies (Brookshire et al., 2024; Recht et al., 2019). In this approach, all images (Section 2.1) collected from a given site were assigned exclusively to one subset (training, validation, or testing). This ensured evaluation on geographically independent sites and prevented the model from learning site-specific characteristics that could artificially improve performance.

*4.1.2 Patch-based sampling strategy*

The preprocessing workflow further employed a patch-based sampling strategy, because the original underwater images possessed large spatial resolutions and frequently contained highly imbalanced regions in which kelp occupied only a small portion of the scene. Training directly on full-resolution images would therefore be computationally inefficient and could result in severe class imbalance, which is a common challenge in semantic segmentation tasks (Chen et al., 2018; Ronneberger et al., 2015). To address this issue, fixed-size patches of 512 × 512 pixels were randomly extracted from each image-mask pair, following patch-based training approaches widely used in remote sensing and medical image segmentation (Isensee et al., 2021; Long et al., 2015).

For each image, up to 20 random patch locations were sampled to increase the number of training samples while preserving spatial diversity. Each extracted patch was evaluated using its corresponding binary mask, and the fraction of kelp pixels within the patch was computed. Patches containing a kelp fraction greater than or equal to 1% were retained as kelp-rich samples, whereas patches containing less than 1% kelp were considered background and retained only with a small probability to maintain class balance. This strategy ensured that the dataset contained sufficient kelp samples for effective model training while still preserving representative background samples.

To maintain the integrity of the site-wise split, patch extraction was performed independently for the training, validation, and test sets using predefined split lists, so that patches generated from images belonging to one split were saved only in the corresponding folder. This prevented patches originating from the same image from appearing in multiple subsets and eliminates the possibility of data leakage, which is known to cause overly optimistic performance estimates in machine learning experiments (Brookshire et al., 2024). Different sampling limits were applied to control the size of each subset, with up to 20 patches per image for the training set and up to 10 patches per image for both validation and test sets. Since kelp typically occupies only a small portion of underwater scenes, the sampling strategy was intentionally biased toward kelp-containing patches while still retaining a subset of background patches to avoid over-prediction of kelp regions. As a result, the training dataset contained substantially more patches than the validation and test datasets and also the validation site remained geographically independent of the training sites.

*4.1.3. Data Augmentation*

To improve model robustness and generalization across varying underwater conditions, data augmentation was applied during training using random geometric and photometric transformations. The augmentation pipeline included horizontal and vertical flips, random rotations, affine transformations, brightness and contrast adjustments, Gaussian noise addition, contrast-limited adaptive histogram equalization (CLAHE), and image blurring. These augmentations were designed to simulate variability in underwater illumination, turbidity, orientation, and imaging conditions commonly encountered in coastal underwater environments.

The proposed preprocessing pipeline was designed to reduce class imbalance between kelp and background, increase the number of effective training samples, preserve spatial and environmental diversity, prevent data leakage between dataset splits, and enable reliable cross-site generalization, which are essential requirements for robust DL models in environmental image analysis (Long, et al., 2015;

Ronneberger, et al., 2015; Torralba and Efros, 2011). This strategy allows the segmentation model to learn site-independent kelp features while providing a realistic and unbiased evaluation on unseen locations.

### 4.2. Deep Learning Architectures

Three deep learning-based semantic segmentation architectures were selected to evaluate different segmentation strategies with varying levels of contextual feature learning. All three architectures were trained using supervised learning with underwater RGB images and their corresponding ground-truth binary segmentation masks generated through manual or SSeg-assisted annotation. The ResNet34-U-Net architecture was chosen as a reliable encoder–decoder framework due to its strong spatial localization capability and efficient training performance (He et al., 2016). The ResNet50-DeepLabV3 architecture was included to evaluate the impact of multi-scale contextual feature extraction using Atrous Spatial Pyramid Pooling (ASPP) (Chen et al., 2018). In addition, a hybrid ResNet50-ASPP-Transformer architecture was developed to examine the role of transformer-based global contextual learning and long-range spatial dependency modeling in underwater kelp segmentation (Dosovitskiy et al., 2021). Together, these architectures provided a comparative analysis of conventional encoder–decoder methods, advanced convolutional-based models, and hybrid convolution–transformer approaches under challenging underwater imaging conditions. All models were trained using 512 × 512 RGB image patches and produced binary segmentation masks for kelp detection.

#### *4.2.1. ResNet34-U-Net architecture*

A ResNet34-U-Net architecture was implemented for semantic segmentation of underwater kelp forests from RGB images (in supplementary Fig. A1). A ResNet34 encoder served as the backbone feature extractor, enabling the model to learn hierarchical spatial representations at multiple scales (He et al., 2016). During feature extraction, the encoder progressively reduced spatial resolution while increasing feature depth to capture high-level semantic information related to kelp structures.

The decoder followed the standard U-Net design and reconstructed high-resolution segmentation maps through progressive upsampling combined with skip connections (Ronneberger et al., 2015). These skip connections transferred fine-grained spatial information from the encoder to the decoder, helping the model preserve object boundaries and improve kelp region reconstruction (Long et al., 2015). Finally, a 1 × 1 convolution layer followed by a sigmoid activation function generated the binary kelp segmentation mask at the original image resolution (Chen et al., 2018).

#### *4.2.2. ResNet50-DeepLabV3 architecture*

A ResNet50-DeepLabV3 architecture was implemented to perform semantic segmentation of underwater kelp forests from RGB imagery. A ResNet50 backbone network was used to extract hierarchical features at multiple spatial levels, allowing the model to learn complex underwater textures and structural patterns effectively (He et al., 2016).

DeepLabV3 employed atrous (dilated) convolution to increase the receptive field without significantly reducing spatial resolution, making it suitable for dense semantic segmentation tasks (Chen et al., 2018). To further improve contextual feature extraction, an Atrous Spatial Pyramid Pooling (ASPP) module was integrated using parallel dilated convolutions with different dilation rates. This approach enabled the model to capture multi-scale contextual information and better identify kelp structures with varying sizes, shapes, and spatial distributions. The architecture (in supplementary Fig. A2) was particularly useful in handling challenging underwater conditions such as illumination changes, turbidity, and complex benthic backgrounds. The final segmentation output was produced using a 1 x 1 convolution layer followed by a sigmoid activation function.

*4.2.3. ASPP-Transformer architecture*

We developed a hybrid ResNet50-ASPP-Transformer architecture for semantic segmentation of underwater kelp forests from RGB imagery (in supplementary Fig. A3). The model utilized a ResNet50 encoder as the backbone network to extract hierarchical spatial and semantic features (He et al., 2016).

To improve multi-scale contextual learning, an Atrous Spatial Pyramid Pooling (ASPP) module was incorporated after the encoder. The ASPP module used parallel dilated convolutions with different dilation rates to capture contextual information at multiple scales (Chen et al., 2018). This helped enhance segmentation performance for kelp structures exhibiting variations in size, shape, and spatial arrangement.

A Transformer block was subsequently integrated to model long-range spatial dependencies and global contextual relationships within the extracted feature maps. Unlike conventional convolution-only architectures, transformer-based attention mechanisms have been shown to capture global feature interactions more effectively, thereby improving segmentation performance in complex underwater environments (Vaswani et al., 2017; Dosovitskiy et al., 2021).

The decoder progressively reconstructed high-resolution segmentation maps through feature upsampling and integration of intermediate encoder features, thereby preserving spatial details and boundary information (Ronneberger et al., 2015). Finally, a 1 × 1 convolution layer followed by a sigmoid activation function was applied to generate the binary kelp segmentation mask at the original image resolution.

## 4.3. Hyperparameter

All models were trained using standardized image subsets (512 × 512 pixels) extracted from the annotated underwater imagery dataset. To ensure a fair comparison among architectures, identical preprocessing, data augmentation, training, and evaluation procedures were applied across all models. During training, model parameters were iteratively adjusted to minimize differences between predicted segmentation masks and manually annotated reference masks, while an independent validation dataset was used to optimize model settings and reduce overfitting. Final binary kelp masks were generated using thresholds selected based on validation performance. This standardized workflow ensured that differences in segmentation accuracy reflected the capabilities of the model architectures rather than inconsistencies in model development or evaluation.

**Table 1.** Hyperparameters used for training the three deep learning models (ResNet34–U-Net, ResNet50–DeepLabV3, and ASPP–Transformer) for kelp segmentation from underwater RGB images.

| Model | Backbone | Batch size | Learning Rate | Key Differences |
|---|---|---|---|---|
| ResNet34–U-Net | ResNet34 | 16 | 1e-4 | Baseline encoder-decoder framework |
| ResNet50–DeepLabV3 | ResNet50 | 8 | 1e-4→5e-5 | Backbone freeze (10 epochs) |
| ASPP–Transformer | ResNet50 | 4 | 1e-5 | Gradient clipping and transformer optimization |

Model-specific training parameters are summarized in Table 1, while the common training settings applied across all models, including optimization methods, loss functions, data augmentation procedures, image normalization, evaluation metrics, threshold selection, and reproducibility settings, are provided in Table A1 (Supp. Table A2).

### 4.4. Performance Metrics

To quantitatively evaluate the performance of the proposed model, several standard evaluation metrics are employed, including Pixel Accuracy, Intersection over Union (IoU), Precision, Recall, F1-score, and Dice Coefficient. Pixel Accuracy measures the proportion of correctly classified pixels over the total number of pixels. It provides an overall assessment of the model's performance but may be biased in the presence of class imbalance

$$Pixel\ Accuracy = \frac{TP+TN}{TP+TN+FP+FN} \quad (1)$$

Where TP, TN, FP, and FN denote true positive, true negative, false positive, and false negative pixels, respectively. Intersection over Union (IoU), also known as the Jaccard Index, evaluates the overlap between the predicted segmentation and the ground truth.

$$IoU = \frac{TP}{TP+FP+FN} \quad (2)$$

Precision measures the proportion of correctly predicted positive pixels among all predicted positive pixels. It reflects the model's ability to minimize false positives.

$$Precision = \frac{TP}{TP+FP} \quad (3)$$

Recall, also known as sensitivity, measures the proportion of actual positive pixels that are correctly identified by the model. It indicates the model's ability to detect relevant regions.

$$Recall = \frac{TP}{TP+FN} \quad (4)$$

The F1-score is the harmonic mean of Precision and Recall, providing a balanced measure of the model's accuracy, especially when dealing with imbalanced datasets.

$$F1 = 2 \times \frac{Precision \times Recall}{Precision + Recall} \quad (5)$$

The Dice Coefficient measures the similarity between the predicted segmentation and the ground truth.

$$Dice = \frac{2TP}{2TP+FP+FN} \quad (6)$$

The equations (1) – (6) are used to evaluate the performance of the proposed model by measuring accuracy, overlap, and classification effectiveness at the pixel level.

## 5. Results

### 5.1. Experimental setup

All experiments were implemented using the PyTorch deep learning framework on a high-performance computing (HPC) cluster managed through the SLURM workload scheduler. Model training was performed on a single NVIDIA A100 GPU with 8 CPU cores and 64 GB RAM in the DGXA100 partition. Patch-based underwater image datasets with a consistent preprocessing and data augmentation were used during training. All three segmentation architectures (Section 4.3) were trained under identical computational and hyperparameter settings (Section 4.4) to ensure fair comparative evaluation.

### 5.2. Model development

The ResNet34-U-Net model exhibited stable convergence behavior with a gradual reduction in both training and validation loss throughout the training process (Fig. 4a and 4b, Supp. Table A3). The validation Dice coefficient increased consistently and stabilized after approximately 40 epochs, indicating effective feature learning and strong generalization capability across underwater scenes. The model achieved the highest validation performance among the developed models, with Dice and IoU values of 0.7927 and 0.6804, respectively. Similarly, the ResNet50-DeepLabV3 architecture demonstrated strong convergence characteristics with consistently decreasing training and validation losses (Fig. 4c, and 4d, Supp. Table A3). The model achieved validation Dice and IoU values of 0.7819 and 0.6661, respectively, demonstrating segmentation performance comparable to the ResNet34-U-Net framework.

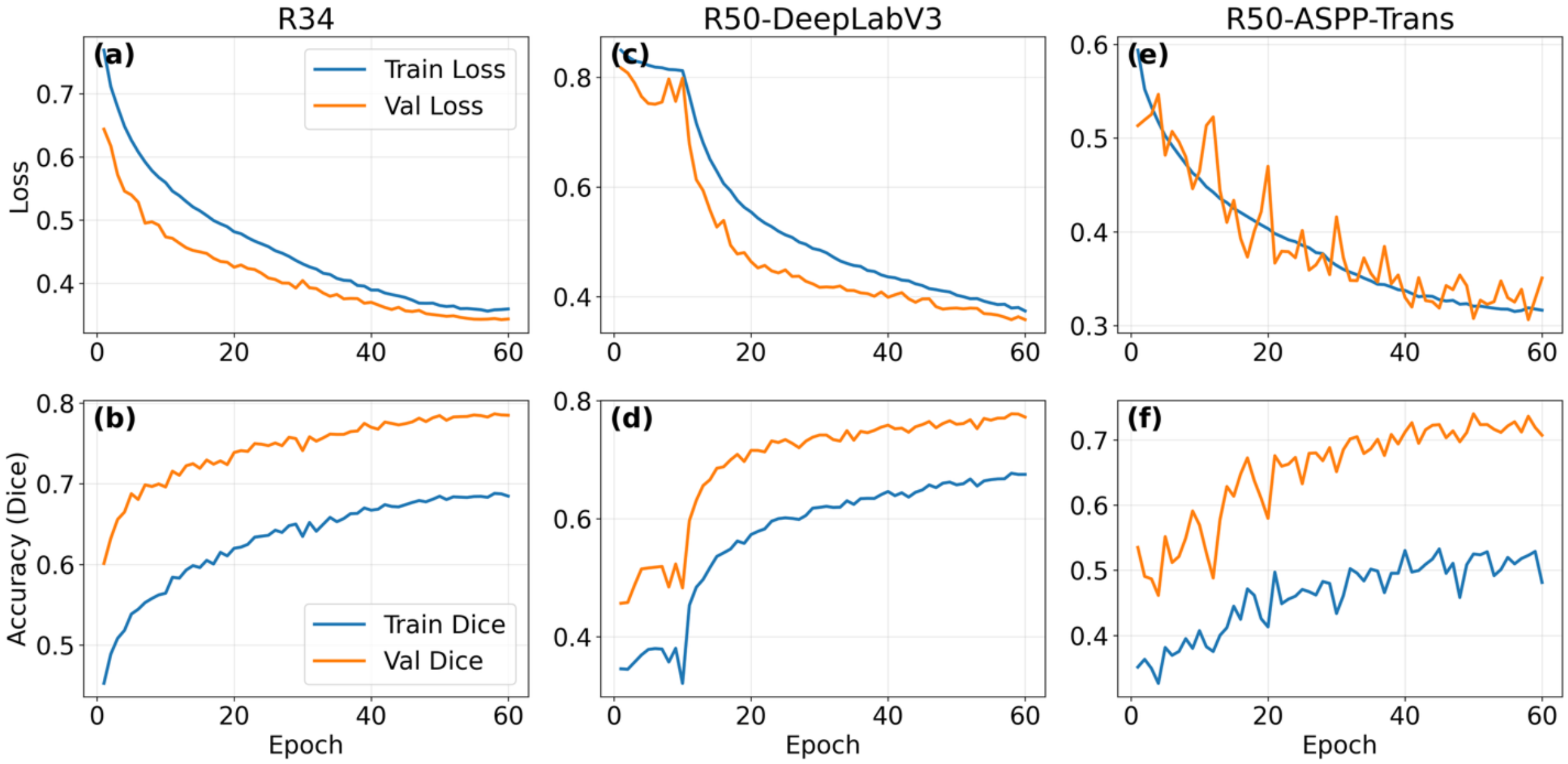

**Figure. 4.** Training and validation loss (top) and Dice accuracy (bottom) for ResNet34-U-Net, ResNet50-DeepLabV3, and ASPP-Transformer models during underwater kelp segmentation model development.

The hybrid ResNet50-ASPP-Transformer model showed larger fluctuations in both training and validation metrics during the early training epochs (Fig. 4e and 4f, Supp. Table A3). However, the model progressively stabilized during the later training epochs and demonstrated effective global contextual feature learning capability for optically complex underwater scenes. The model achieved relatively lower quantitative performance compared with the other two models, with validation Dice and IoU values of 0.7480 and 0.6275, respectively. Overall, the training results indicate that all three models achieved effective convergence and demonstrated reliable underwater kelp segmentation capability without significant overfitting. Among the developed models, ResNet34-U-Net and DeepLabV3 exhibited comparatively more stable and consistent throughout the training process than the hybrid ResNet50-ASPP-Transformer model, indicating stronger optimization stability and generalization under this training condition.

### 5.3. Evaluation on independent test sites

Among the models, ResNet50-DeepLabV3 performed the best segmentation performance on independent test data (Table 2). The independent test dataset consisted of 1,556 annotated image patches (512 × 512 pixels) generated from underwater images collected at the held-out test sites, Bald Porcupine and Ram Island (Supplementary Table 1), which were excluded from all training and validation procedures. It had Dice and IoU values of 0.7120 and 0.6267, respectively. The ResNet34-U-Net architecture also demonstrated strong segmentation performance, achieving Dice and IoU values of 0.6868 and 0.5978, respectively, indicating effective boundary reconstruction and stable generalization capability across unseen underwater environments. The hybrid ASPP-Transformer model achieved the highest pixel accuracy (0.8528) among the developed models, however, its Dice and IoU values were comparatively lower than those of the convolution-based models.

**Table 2.** Quantitative evaluation of three semantic segmentation models for underwater kelp detection on the test dataset.

| Model | Pixel Acc | IoU | Precision | Recall | F1-score | Dice | N (Patches) |
|---|---|---|---|---|---|---|---|
| ResNet34-U-Net | 0.8327 | 0.5978 | 0.6392 | 0.8469 | 0.6868 | 0.6868 | 1556 |
| ResNet50-DeepLabV3 | 0.8408 | 0.6267 | 0.6625 | 0.8678 | 0.7120 | 0.7120 | |
| ASPP-Transformer | 0.8528 | 0.5746 | 0.6588 | 0.7167 | 0.6437 | 0.6437 | |

Figures 5 and 6 present qualitative comparisons of kelp segmentation results obtained from the three models using independent underwater imagery collected from the Bald Porcupine and Ram Island sites, respectively. In particular, Figure 6 shows the models' ability to discriminate between kelp and seagrass at Ram Island. Additional qualitative results are presented in Supplementary Figures A4–A6. The visual comparisons include the original RGB imagery, SSeg-assisted ground-truth masks, and predicted segmentation outputs together with estimated kelp cover percentages and pixel-wise accuracy statistics. Overall, the ResNet34-U-Net and ResNet50-DeepLabV3 architectures produced segmentation outputs that more closely matched the ground-truth masks, particularly in regions with complex kelp morphology, overlapping fronds, and heterogeneous benthic backgrounds. The DeepLabV3 framework demonstrated improved boundary delineation and more consistent segmentation performance under varying underwater conditions. In contrast, the ASPP-Transformer model occasionally produced fragmented segmentation outputs in optically complex underwater scenes. Overall, the qualitative evaluation highlights the importance of robust multi-scale feature extraction and accurate boundary reconstruction for reliable underwater kelp segmentation.

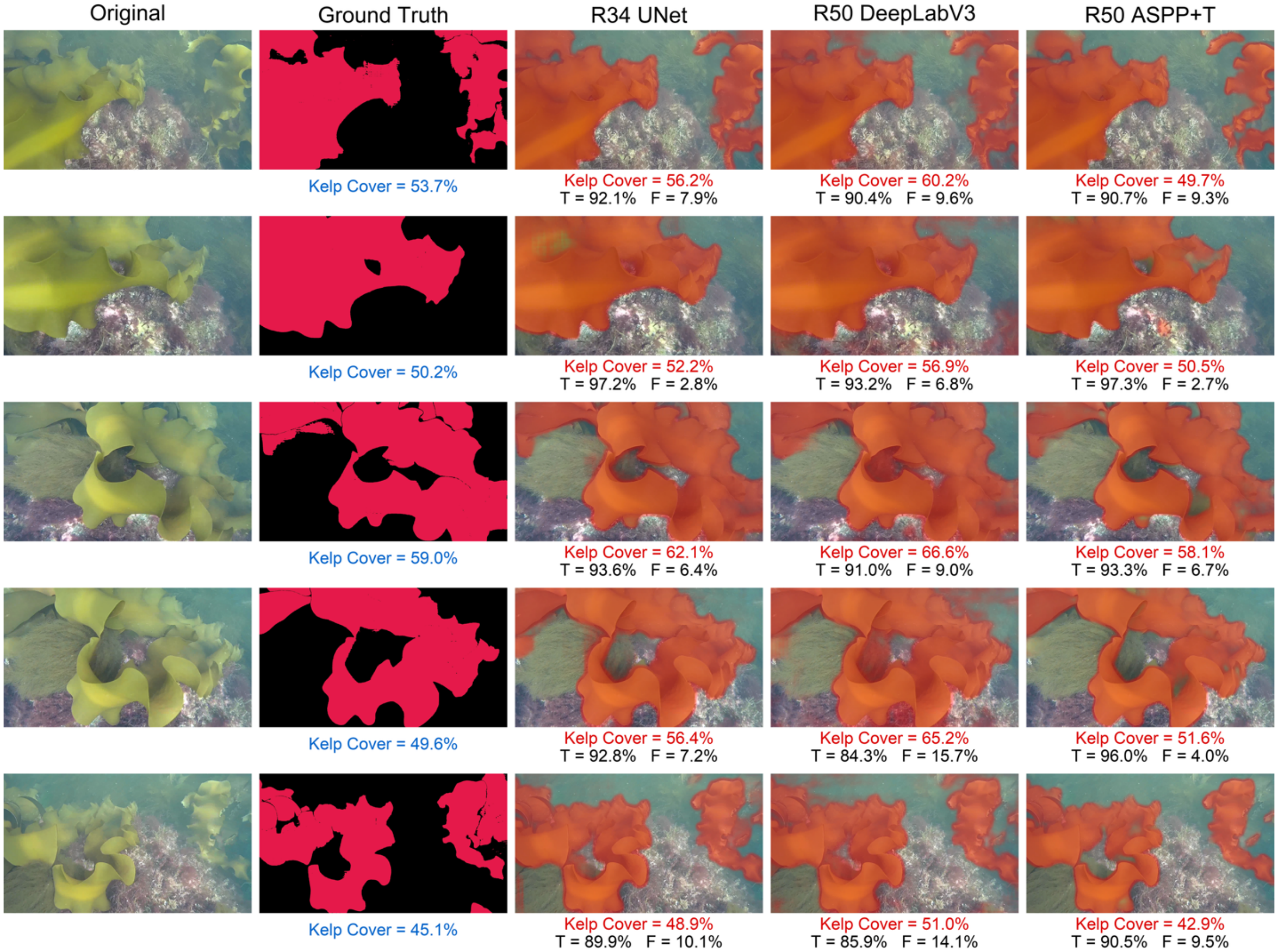

**Figure 5.** Qualitative comparison of kelp segmentation results obtained using three DL models (ResNet34-U-Net, ResNet50-DeepLabV3, and ASPP-Transformer) together with original images and ground-truth masks for unseen underwater images acquired at the independent Bald Porcupine site. Predicted kelp cover and pixel-wise accuracy statistics are displayed for each model.

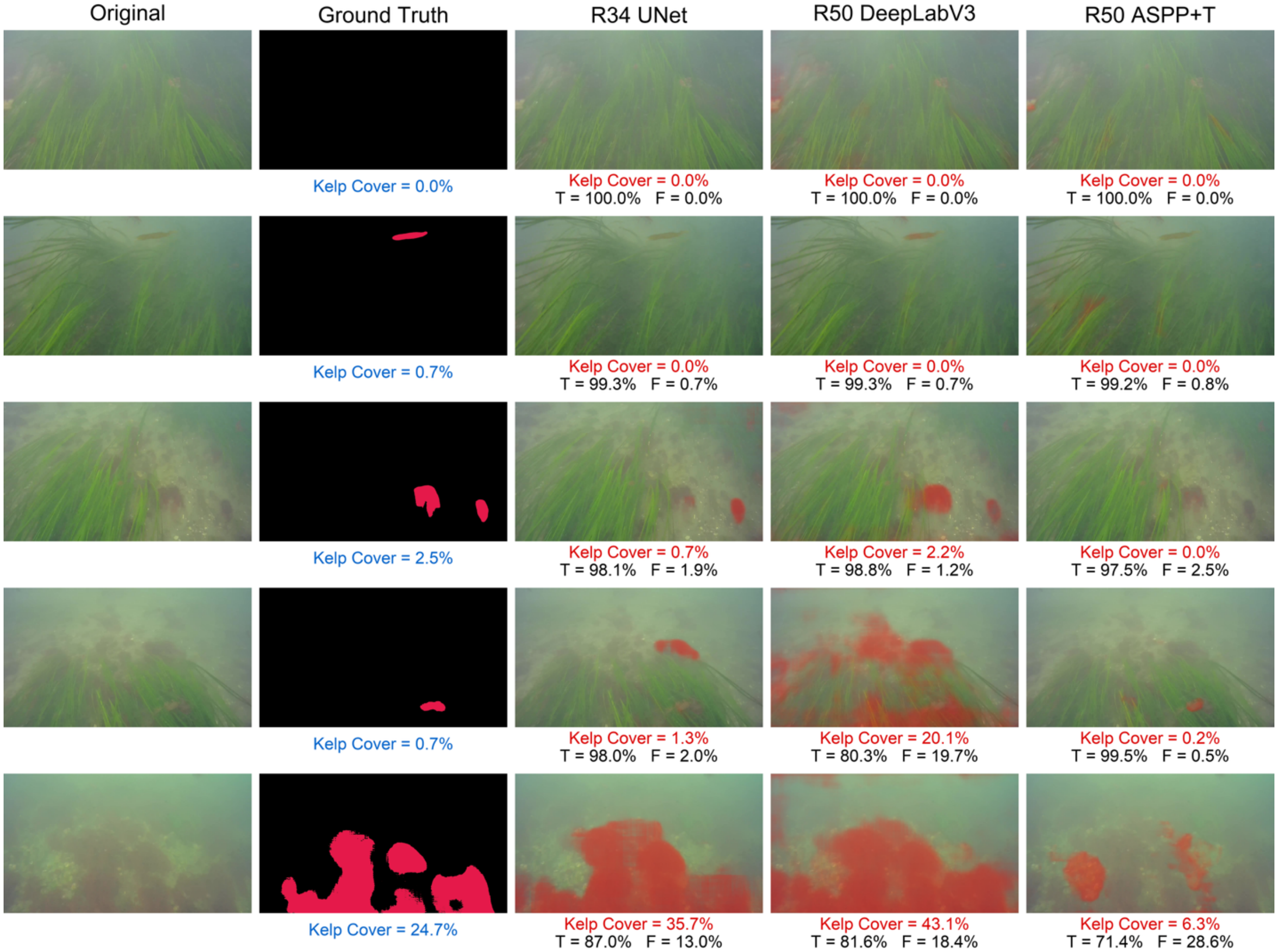

**Figure 6.** Qualitative comparison of kelp segmentation results obtained using three DL models (ResNet34-U-Net, ResNet50-DeepLabV3, and ASPP-Transformer) together with original images and ground-truth masks for unseen underwater images acquired at the independent Ram Island site. Predicted kelp cover and pixel-wise accuracy statistics are displayed for each model. Note that these images also contain seagrass, as we were interested in how well the model discriminated between seagrass and kelp.

### 5.4. Generalization to unseen survey sites

Unlike the independent test sites used in Section 5.3, the sites evaluated here were not included in the development dataset and were used to assess model generalization across additional kelp habitats. Overall, the ResNet50-DeepLabV3 model achieved the best and most consistent segmentation performance across the majority of sites (Fig. 7). Dice scores for DeepLabV3 generally ranged between approximately 35–82%, while IoU values varied from about 20–70%. The highest segmentation performance was observed at Sites 2, 9, 10, and 15, where Dice values exceeded ~70% and IoU values reached ~60–70%, indicating strong agreement with the ground-truth masks. In comparison, the ResNet34-U-Net model produced relatively stable performance across sites, with Dice values mostly ranging from ~40–80% and IoU values between ~25–65%.

The R50 ASPP-Transformer model showed larger variability across sites. Although the model achieved competitive performance for certain sites, such as Sites 6, 10, and 13, where Dice values approached ~70–75%, performance declined substantially for several other locations. For example, Dice values dropped below ~20% at Sites 18 and 19, with corresponding IoU values near ~5–10%, indicating poor segmentation performance under challenging underwater conditions. These reductions likely resulted from variations in illumination, water clarity, shadows, complex benthic backgrounds, and sparse kelp distribution.

The kelp cover comparison further demonstrates the ability of the models to reproduce site-level spatial variability in kelp abundance. Ground-truth kelp cover varied considerably among sites, ranging from near 0% at Sites 18 and 19 to approximately 45–50% at Sites 3 and 13. The DeepLabV3 and ResNet34-U-Net models generally followed similar trends to the ground-truth observations across most sites, although some overestimation and underestimation were observed. For example, at Site 3, where the ground-truth kelp cover was approximately 50%, all models underestimated kelp cover to varying degrees. Conversely, at several moderate-cover sites, predicted kelp cover values closely matched the ground truth, indicating robust model generalization.

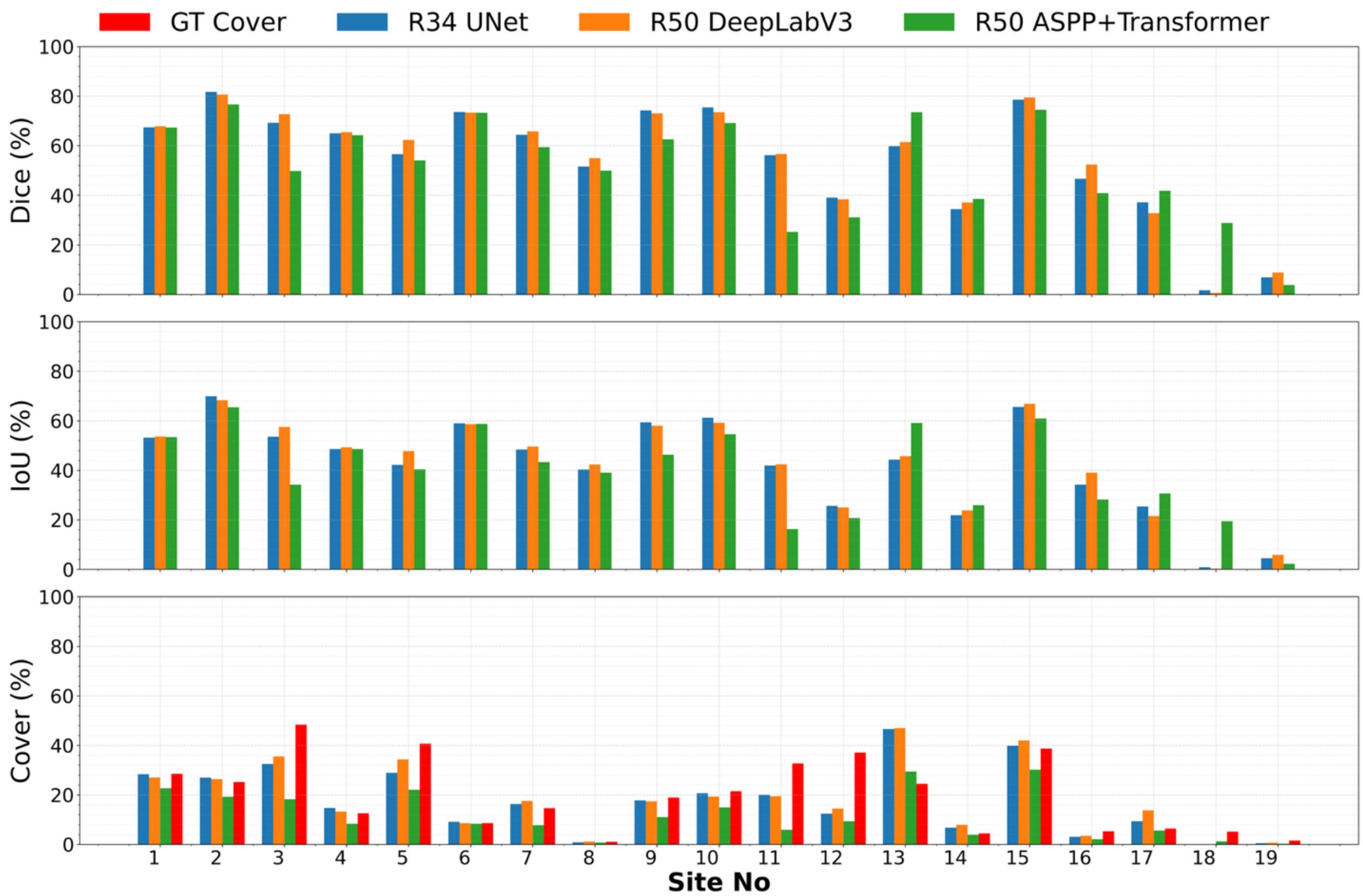

**Figure 7.** Site-wise comparison of segmentation performance metrics, including Dice (%) and IoU (%), together with estimated kelp cover (%) derived from three deep learning–based semantic segmentation models and corresponding ground-truth kelp cover across 19 independent study sites. For each site, 50 underwater images were analyzed, kelp cover was calculated individually for all images, and the median kelp cover value was used to represent the final site-level estimate for each model.

### 5.5. Transferability to external Data Sets

To further evaluate model transferability, the trained models were applied to the external DFO (Charlotte and Woody Island) and New Hampshire datasets (Section 2.3), both of which were acquired independently of this study and excluded from all stages of model development. Figure 8 shows representative segmentation results and corresponding estimated kelp cover (%) for the external datasets. All three models successfully detected the dominant kelp beds across the DFO and New Hampshire imagery; however, substantial differences were observed in the estimated kelp cover. For the Woody Island examples, all models underestimated kelp cover relative to the ground truth, with estimated cover ranging from 9.8–15.4% for a ground-truth cover of 27.8% and from 15.2–33.4% for a ground-truth cover of 51.8%. Similar variability was observed for the Charlotte Island examples, where DeepLabV3 produced considerably lower kelp cover estimates (21.8–50.2%) than ResNet34-U-Net (55.2–66.7%) for ground-truth covers of 62.8% and 70.1%. In contrast, all models showed closer agreement with the New Hampshire example, with estimated kelp cover ranging from 76.6% to 81.7% compared to a ground-truth cover of 82.3%. Overall, ResNet34-U-Net generally produced the highest kelp cover estimates, whereas the ASPP-Transformer model yielded the most conservative predictions. DeepLabV3 showed variable behavior, producing both the lowest and highest cover estimates depending on image characteristics and site conditions.

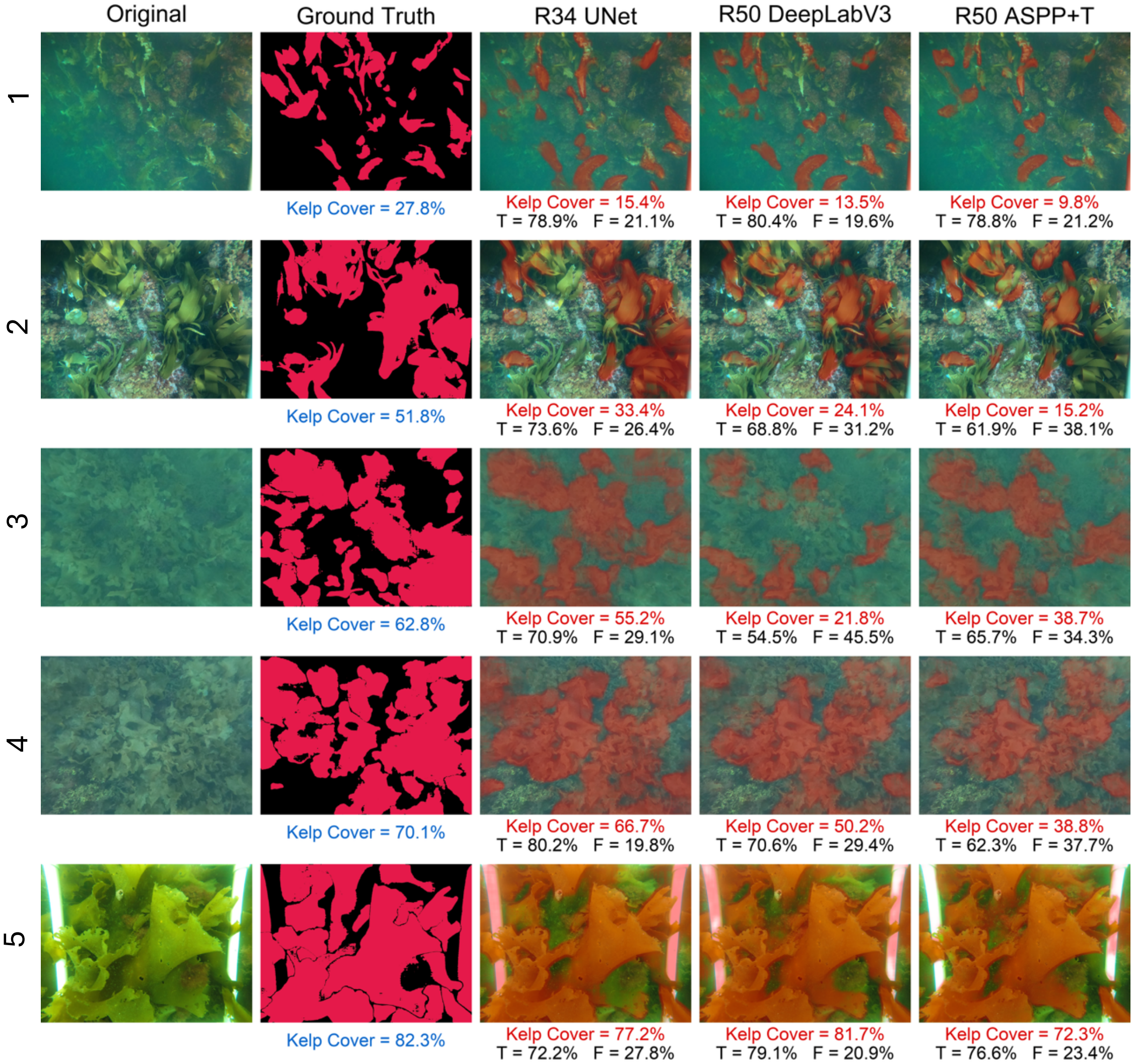

**Figure 8.** Qualitative comparison of kelp segmentation results obtained using three deep learning–based semantic segmentation models (ResNet34 U-Net, ResNet50 DeepLabV3, and ResNet50 ASPP+Transformer) for underwater images acquired from the external datasets from Woody Island (row 1&2), Charlotte (row 3 & 4), and New Hampshire (row 5).

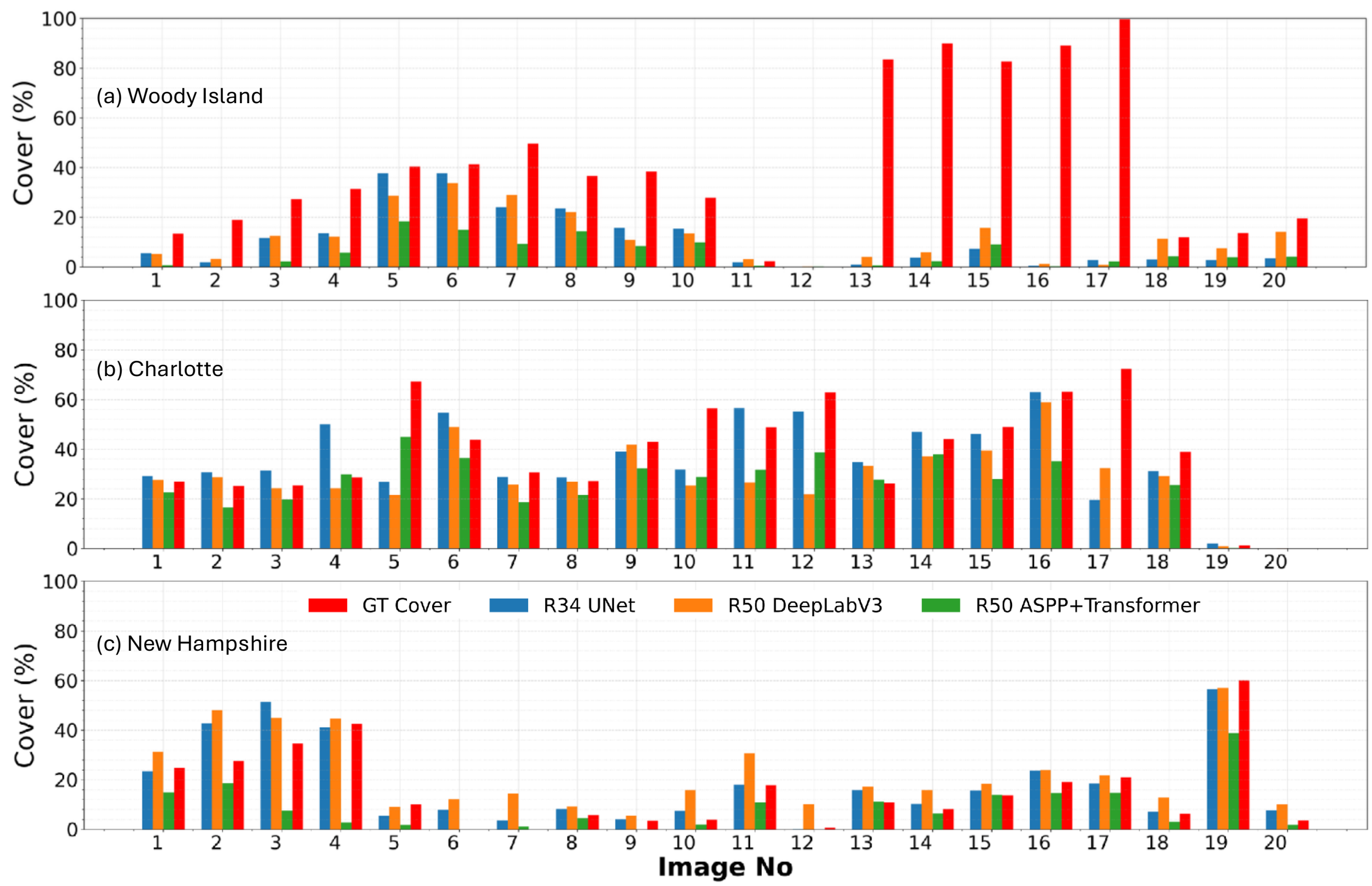

**Figure 9.** Image-wise comparison of kelp cover (%) derived from ground-truth masks and three deep learning-based semantic segmentation models (ResNet34-U-Net, ResNet50-DeepLabV3, and ResNet50-ASPP-Transformer) for 20 representative underwater images from the external datasets (a), Woody Island, (b). Charlotte, and (c). New Hampshire coastal waters.

Along with the qualitative image assessment, Figure 9 compares image-wise kelp cover (%) derived from ground-truth masks and model predictions for 20 representative images from the external sites such as Woody Island, Charlotte, and New Hampshire datasets. At Woody Island, ground-truth kelp cover ranged from approximately 2–100%, whereas model-derived estimates were substantially lower, with ResNet34-U-Net, DeepLabV3, and ASPP-Transformer generally predicting cover values below 40%, 35%, and 25%, respectively. The largest discrepancies occurred for images with dense kelp cover (>80%), where all models underestimated kelp extent. In contrast, the Charlotte dataset showed closer agreement with ground-truth cover, which ranged from approximately 2–73%. ResNet34-U-Net produced predictions ranging from approximately 20–63%, closely tracking the observed variability, while DeepLabV3 and ASPP-Transformer typically generated lower cover estimates, particularly for images with moderate to high kelp abundance. For the New Hampshire dataset, ground-truth kelp cover ranged from approximately 0–60%, and all three models showed relatively good agreement with the reference values. Predicted kelp cover generally ranged from 0–57% for ResNet34-U-Net, 0–58% for DeepLabV3, and 0–40% for ASPP-Transformer. Overall, ResNet34-U-Net consistently produced the highest kelp cover estimates, ASPP-Transformer the most conservative estimates, and DeepLabV3 intermediate values across the external datasets.

## 6. Discussion

Recent advances in computer vision and deep learning have enabled efficient and scalable quantification of kelp forests from underwater imagery, supporting ecological monitoring and management. Among the evaluated models, ResNet50-DeepLabV3 consistently achieved the strongest overall performance across the development, independent test, and external datasets and is therefore recommended for automated nearshore subsurface kelp segmentation. Its Atrous Spatial Pyramid Pooling module effectively captured multi-scale features, improving robustness under varying underwater optical conditions and complex benthic environments and contributing to the highest Dice and IoU scores among the evaluated models. The model also performed well in delineating fragmented kelp patches and fine-scale boundaries. In contrast, ResNet34-U-Net frequently produced broader segmentation outputs and higher kelp cover estimates in optically complex scenes, suggesting increased segmentation sensitivity. Although the ASPP-Transformer architecture achieved high pixel accuracy, its lower Dice and IoU values suggest that the available training dataset may not have been sufficiently large to fully realize the advantages of transformer-based feature learning.

Underwater optical variability, environmental complexity, and domain shift influenced segmentation performance and model generalization across study sites. Because the training dataset was derived primarily from northeastern U.S. coastal waters, it may not fully represent the broader variability of kelp ecosystems worldwide. Differences in kelp morphology, species composition, water clarity, illumination, and image acquisition conditions can affect model performance when applied to new environments. These effects were evident in the external validation datasets, where model predictions varied among sites and environmental conditions. In addition, seasonal and long-term ecological variability were not examined and may further influence model transferability. These findings highlight the importance of independent validation and geographically diverse training datasets for underwater kelp segmentation.

A key component of this study was the development of a semi-automated annotation workflow using SSeg to generate pixel-level kelp masks from underwater imagery. This approach substantially reduced annotation effort while enabling the creation of a large training dataset of 3,395 image–mask pairs. The semi-annotated workflow demonstrates a practical strategy for rapidly generating training data while maintaining annotation quality through targeted manual review. However, complex underwater scenes containing overlapping kelp, variable illumination, motion blur, and visually similar benthic features often required manual refinement. Because annotation quality directly influences model performance, future efforts should focus on standardized annotation protocols, quality-control procedures, and active-learning approaches to improve efficiency, reduce labeling bias, and support the development of larger and more diverse underwater imagery datasets.

Evaluation on the independent and external datasets demonstrated that the trained models could identify kelp across a range of underwater environments beyond those represented in the development dataset. Among the evaluated architectures, ResNet50-DeepLabV3 provided the most reliable balance between segmentation accuracy and kelp cover estimation. While performance varied with environmental conditions, the results suggest that DeepLabV3 offers a robust framework for automated kelp mapping. For future applications, we recommend validating predictions using a representative subset of annotated images and, where possible, fine-tuning the model with local training data. Expanding training datasets to include broader geographic regions, environmental conditions, and kelp morphologies will further improve model transferability and reduce domain-shift effects.

To support broader community adoption, we provide the best-performing model weights and inference workflow as an open tool, Kelp-O-Tron, for automated kelp segmentation from underwater imagery. Kelp-O-Tron enables researchers, monitoring programs, and coastal managers to generate pixel-level kelp masks and kelp cover estimates from new image datasets. Because performance may vary under novel environmental conditions, users are encouraged to validate predictions using a small, annotated subset of their imagery before large-scale application. Detailed installation, inference, and validation instructions are provided in the Supplementary Materials, facilitating broader adoption of deep-learning-based kelp monitoring workflows.

## 7. Conclusion

This study presented a comparative evaluation of three deep learning–based semantic segmentation architectures for underwater kelp detection using high-resolution underwater RGB imagery acquired across diverse coastal environments. A large-scale underwater image dataset with SSeg-assisted segmentation masks was developed to train and evaluate the models under varying underwater optical conditions. The results demonstrated that all evaluated architectures could detect underwater kelp with good segmentation accuracy despite challenges associated with illumination variability, turbidity, complex benthic backgrounds, and domain heterogeneity.

Among the evaluated frameworks, ResNet50-DeepLabV3 achieved the most balanced and consistent performance across independent test sites, producing higher Dice and IoU scores together with comparatively stable kelp cover estimation. Independent evaluation across external sites such as Woody Island, Charlotte, and New Hampshire, further demonstrated good model generalization, although variability in kelp cover predictions highlighted the continuing influence of underwater optical complexity and domain shift. Future improvements could be achieved through larger geographically diverse training datasets and inclusion of additional kelp morphologies and environmental conditions.

Overall, the study demonstrates that deep learning–based semantic segmentation provides a promising framework for automated subsurface kelp mapping and ecological habitat assessment. Further, we provide tools for other users to either use this model as is (https://github.com/Sundarabalan/KelpSegModels) or to retrain them for your own system. Finally, the integration of underwater segmentation products with future drone-based and satellite-based remote sensing observations could support scalable kelp monitoring, marine habitat characterization, and long-term coastal ecosystem management.

**Acknowledgements:**

We acknowledge Claudio DiBacco for co-leading the research that led to image acquisition in Nova Scotia.

**Funding:**

This research was funded under the NASA MUREP OCEAN Grant, number 80NSSC21M0358 supporting SB, LNW, MLM, and JEKB and grant AIA2025-163563-C31 supporting CB and ACM. Additionally, LNW was supported by a Healthy Estuary Grant from the Massachusetts Bays National Estuary Partnership under EPA Grant Award 00A015460CE, and SB received additional funding from the Massachusetts Department of Marine Fisheries. Funding for image collection in Nova Scotia, Canada was provided by Fisheries and Oceans Canada.

## References

- Akkaynak, D., and Treibitz, T., 2019. Sea-thru: A method for removing water from underwater images. In: Proceedings of the IEEE/CVF Conference on Computer Vision and Pattern Recognition (CVPR), pp. 1682–1691.

- Alonso, I., Sabater, A., Ferstl, D., Montesano, L., Murillo, A. C., 2021. Semi-Supervised Semantic Segmentation With Pixel-Level Contrastive Learning From a Class-Wise Memory Bank, Proceedings of the IEEE/CVF International Conference on Computer Vision (ICCV), pp. 8219-8228
- Anwar, S., and Li, C., 2020. Diving deeper into underwater image enhancement: A survey. Signal Processing: Image Communication 89, 115978. https://doi.org/10.1016/j.image.2020.115978
- Bartsch, I., Wiencke, C., Bischof, K., Buchholz, C.M., Buck, B.H., Eggert, A., Feuerpfeil, P., Hanelt, D., Jacobsen, S., Karez, R., Karsten, U., Molis, M., Roleda, M.Y., Schubert, H., Schumann, R., Valentin, K., Weinberger, F. and Wiese, J., 2008. The genus *Laminaria* sensu lato: recent insights and developments. *European Journal of Phycology*, 43(1), pp.1–86
- Bell, T.W., Allen, J.G., Cavanaugh, K.C., Siegel, D.A., 2020. Three decades of variability in California's giant kelp forests from the Landsat satellite record. Remote Sensing of Environment 238, 110811.
- Bell, T.W., Cavanaugh, K.C., Saccomanno, V.R., Houskeeper, H.F., Eddy, N., et al. (2023) Kelpwatch: A new visualization and analysis tool to explore kelp canopy dynamics reveals variable response to and recovery from marine heatwaves. PLoS ONE 18(3): e0271477.
- Borja, C., Plou, C., Martinez-Cantín, R., Murillo, A.C., 2026. SSeg: Active Sparse Point-Label Augmentation for Semantic Segmentation. In: Proceedings of the IEEE/CVF Winter Conference on Applications of Computer Vision Workshops (WACVW 2026), CV4EO Workshop.
- Borowiec, M.L., Dikow, R.B., Frandsen, P.B., McKeeken, A., Valentini, G., White, A.E., 2022. Deep learning as a tool for ecology and evolution. *Methods Ecol. Evol.* 13, 1640–1660. https://doi.org/10.1111/2041-210X.13901.
- Brookshire, G., Kasper, J., Blauch, N.M., Wu, Y.C., Glatt, R., Merrill, D.A., Gerrol, S., Yoder, K.J., Quirk, C., Lucero, C., 2024. Data leakage in deep learning studies of translational EEG. Frontiers in Neuroscience 18, 1373515.
- Byrnes, J.E., Reed, D.C., Cardinale, B.J., Cavanaugh, K.C., Holbrook, S.J. and Schmitt, R.J., 2011. Climate-driven increases in storm frequency simplify kelp forest food webs. *Global Change Biology*, 17(8), pp.2513–2524. https://doi.org/10.1111/j.1365-2486.2011.02409.x
- Cavanaugh, K.C., Siegel, D.A., Reed, D.C., Dennison, P.E., 2011. Environmental controls of giant-kelp biomass in the Santa Barbara Channel, California. Marine Ecology Progress Series 429, 1–17.
- Cavanaugh KC, Siegel DA, Kinlan BP, Reed DC (2010) Scaling giant kelp field measurements to regional scales using satellite observations. Mar Ecol Prog Ser 403:13-27 https://doi.org/10.3354/meps08467
- Chen, L.-C., Zhu, Y., Papandreou, G., Schroff, F., Adam, H., 2018. Encoder–decoder with atrous separable convolution for semantic image segmentation. In: Proceedings of the European Conference on Computer Vision (ECCV), pp. 801–818.
- Dayton, P.K., 1985. Ecology of kelp communities. *Annual Review of Ecology and Systematics*, 16, pp.215–245.
- Dijkstra, J.A., Harris, L.G., Mello, K., Litterer, A., Wells, C., Ware, C., 2017. Invasive seaweeds transform habitat structure and increase biodiversity of associated species. *J. Ecol.* 105, 1668–1678.
- Dosovitskiy, A., Beyer, L., Kolesnikov, A., Weissenborn, D., Zhai, X., Unterthiner, T., Dehghani, M., Minderer, M., Heigold, G., Gelly, S., Uszkoreit, J., Houlsby, N., 2021. An image is worth 16×16 words: Transformers for image recognition at scale. In: International Conference on Learning Representations (ICLR).
- Durden, J.M., Schoening, T., Althaus, F., Friedman, A., Garcia, R., Glover, A.G., Greinert, J., Stout, N.J., Jones, D.O.B., Jordt, A., Kaeli, J.W., et al., 2016. Perspectives in visual imaging for marine biology and ecology: From acquisition to understanding. Oceanography and Marine Biology: An Annual Review 54, 1–72.
- Eger, A.M., Wood, G.V. and Byrnes, J. (2025), An Environmental Niche Exploration Tool for Kelp Forest Management. Ecol Evol, 15: e71459. https://doi.org/10.1002/ece3.71459

- Eger, A.M., Marzinelli, E.M., Bauman, A.G., et al., 2022. Global kelp forest restoration: past lessons, present status, and future directions. Biological Reviews 97, 1449–1475.
- Er, M.J., Chen, J., Zhang, Y., Gao, W., 2023. Research challenges, recent advances, and popular datasets in deep learning-based underwater marine object detection: A review. Sensors 23(4), 1990.
- Filbee-Dexter, K., Wernberg, T., Grace, S. P., Thormar, J., et al., 2020. Marine heatwaves and the collapse of marginal North Atlantic kelp forests. Scientific Reports 10, 13388.
- Finger, D.J., McPherson, M. L., Houskeeper, H. F., and R. M. Kudela, 2021. Mapping bull kelp canopy in northern California using Landsat to enable long-term monitoring. Remote Sensing of Environment 254, 112243.
- Game, C.A., Piechaud, N., Howell, K.L., 2026. Deep blueprint: A literature review and guide to automated image classification for ecologists. *J. Anim. Ecol.* 1–28. https://doi.org/10.1111/1365-2656.70271.
- González-Rivero, M., Beijbom, O., Rodriguez-Ramirez, A., Bryant, D.E.P., Ganase, A., Gonzalez-Marrero, Y., Herrera-Reveles, A., Kennedy, E.V., Kim, C.J.S., Lopez-Marcano, S., Markey, K., Neal, B.P., Razak, T.B., Vercelloni, J., Hoegh-Guldberg, O., 2020. Monitoring of coral reefs using artificial intelligence: A feasible and cost-effective approach. *Remote Sensing* 12 (3), 489. https://doi.org/10.3390/rs12030489
- Graham, M.H., Vasquez, J.A., Buschmann, A.H., 2007. Global ecology of the giant kelp Macrocystis: From ecotypes to ecosystems. Oceanography and Marine Biology: An Annual Review 45, 39–88.
- Hamilton, S.L., Bell, T.W., Watson, J.R., Grorud-Colvert, K.A., Menge, B.A., 2020. Remote sensing: Generation of long-term kelp bed datasets for evaluation of impacts of climatic variation. Ecology 101(7), e03031.
- He, K., Zhang, X., Ren, S., Sun, J., 2016. Deep residual learning for image recognition. In: *Proceedings of the IEEE Conference on Computer Vision and Pattern Recognition (CVPR)*, pp. 770–778.
- Hong, Y., Zhou, X., Hua, R., Lv, Q., Dong, J., 2024. WaterSAM: Adapting SAM for underwater object segmentation. Journal of Marine Science and Engineering 12, 1616. https://doi.org/10.3390/jmse12091616
- Isensee, F., Jaeger, P.F., Kohl, S.A.A., Petersen, J., Maier-Hein, K.H., 2021. nnU-Net: a self-configuring method for deep learning-based biomedical image segmentation. Nature Methods 18, 203–211. https://doi.org/10.1038/s41592-020-01008-z
- Jaffe, J.S., 2015. Underwater optical imaging: The past, the present, and the prospects. IEEE Journal of Oceanic Engineering 40(3), 683–700.
- Krause-Jensen, D., and Duarte, C.M. 2016. Substantial role of macroalgae in marine carbon sequestration. Nat. Geosci., 9 (10), pp. 737-742
- Krumhansl, K.A., Brooks, C.M., Lowen, J.B., O'Brien, J.M., Wong, M.C., DiBacco, C., 2024. Loss, resilience and recovery of kelp forests in a region of rapid ocean warming. *Ann. Bot.* 133, 73–92. https://doi.org/10.1093/aob/mcad170.
- Krumhansl, K.A., Okamoto, D.K., Rassweiler, A., Novak, M., et al., 2016 Global patterns of kelp forest change over the past half-century, *Proc. Natl. Acad. Sci. U.S.A.* 113 (48) 13785-13790, https://doi.org/10.1073/pnas.1606102113.
- Li, C., Guo, J., Guo, C., 2018. Emerging from water: Underwater image color correction based on weakly supervised color transfer. IEEE Signal Processing Letters 25 (3), 323–327.
- Li, J., Cai, Y., Li, Q., Kou, M., Zhang, T., 2024. A review of remote sensing image segmentation by deep learning methods. International Journal of Digital Earth 17(1). https://doi.org/10.1080/17538947.2024.2328827
- Long, J., Shelhamer, E., and Darrell, T., 2015. Fully convolutional networks for semantic segmentation. In: Proceedings of the IEEE Conference on Computer Vision and Pattern Recognition (CVPR), pp. 3431–3440.

- Mahmood, A., Ospina, A.G., Bennamoun, M., An, S., Sohel, F., Boussaid, F., Hovey, R., Fisher, R.B., Kendrick, G.A., 2020. Automatic hierarchical classification of kelps using deep residual features. Sensors 20, 447.
- Mann K. H. 1973. Seaweeds: Their Productivity and Strategy for Growth. *Science* **182**, 975-981. DOI:10.1126/science.182.4116.975
- McHenry, J., Okamoto, D.K., Filbee-Dexter, K., et al., 2025. A blueprint for national assessments of the blue carbon capacity of kelp forests applied to Canada's coastline. npj Ocean Sustainability 4, 30.
- Mora-Soto, A.; Palacios, M.; Macaya, E.C.; Gómez, I.; Huovinen, P.; Pérez-Matus, A.; Young, M.; Golding, N.; Toro, M.; Yaqub, M.; et al. A High-Resolution Global Map of Giant Kelp (*Macrocystis pyrifera*) Forests and Intertidal Green Algae (*Ulvophyceae*) with Sentinel-2 Imagery. *Remote Sens.* 2020, *12*, 694. https://doi.org/10.3390/rs12040694
- Noman, M.K., Islam, S.M.S., Abu-Khalaf, J., Jalali, S.M.J., Lavery, P., 2023. Improving accuracy and efficiency in seagrass detection using state-of-the-art AI techniques. *Ecol. Inform.* 76, 102047. https://doi.org/10.1016/j.ecoinf.2023.102047
- Pierce, J., Butler IV, M.J., Rzhanov, Y., Lowell, K., Dijkstra, J.A., 2021. Classifying 3-D models of coral reefs using structure-from-motion and multi-view semantic segmentation. *Front. Mar. Sci.* 8.
- Pierce, J.P., Rzhanov, Y., Lowell, K., Dijkstra, J.A., 2020. Reducing annotation times: Semantic segmentation of coral reef survey images. In: *Global Oceans 2020: Singapore – U.S. Gulf Coast*. IEEE. https://doi.org/10.1109/IEEECONF38699.2020.9389163.
- Pollock, L.J., Kitzes, J., Beery, S., et al., 2025. Harnessing artificial intelligence to fill global shortfalls in biodiversity knowledge. *Nat. Rev. Biodivers.* 1, 166–182. https://doi.org/10.1038/s44358-025-00022-3.
- Raine, S., Marchant, R., Kusy, B., Maire, F., Fischer, T., 2022. Point label aware superpixels for multi-species segmentation of underwater imagery. *IEEE Robot. Autom. Lett.* 7, 8291–8298. https://doi.org/10.1109/LRA.2022.3187836.
- Recht, B., Roelofs, R., Schmidt, L., Shankar, V., 2019. Do ImageNet classifiers generalize to ImageNet? In: Proceedings of the 36th International Conference on Machine Learning (ICML), PMLR 97, 5389–5400.
- Ronneberger, O., Fischer, P., and Brox, T., 2015. U-Net: Convolutional networks for biomedical image segmentation. In: Medical Image Computing and Computer-Assisted Intervention (MICCAI), LNCS, pp. 234–241.
- Saccomanno, V.R., Bell, T., Pawlak, C., Stanley, C.K., Cavanaugh, K.C., Hohman, R., Klausmeyer, K.R., Nickels, A., Hewerdine, W., Garza, C., Fleener, G., Gleason, M., 2023. Using unoccupied aerial vehicles to map and monitor changes in emergent kelp canopy after an ecological regime shift. Remote Sensing in Ecology and Conservation 9, 62–75.
- Sato, M., Kinoshita, J., Ishita, K., et al., 2025. Rapid loss of temperate kelp forests revealed by unmanned aerial vehicle (UAV) photography and underwater observations. Aquatic Botany 200, 103900.
- Schechner, Y.Y., and Karpel, N., 2005. Recovery of underwater visibility and structure by polarization analysis. IEEE Journal of Oceanic Engineering 30(3), 570–587.
- Schroeder, S.B., Boyer, L., Juanes, F., Costa, M., 2020. Spatial and temporal persistence of nearshore kelp beds on the west coast of British Columbia, Canada using satellite remote sensing. Remote Sensing in Ecology and Conservation 6, 327–343.
- Steneck, R.S., Graham, M.H., Bourque, B.J., Corbett, D., Erlandson, J,M., Estes. J.A., and Tenner, M.J. 2002. Kelp forest ecosystems: biodiversity, stability, resilience and future. Environmental Conservation. 29(4):436-459. doi:10.1017/S0376892902000322
- Teagle, H., Hawkins, S.J., Moore, P.J., Smale, D.A., 2017. The role of kelp species as biogenic habitat formers in coastal marine ecosystems. Journal of Experimental Marine Biology and Ecology 492, 81–98. https://doi.org/10.1016/j.jembe.2017.01.017

- Torralba, A., Efros, A.A., 2011. Unbiased look at dataset bias. In: Proceedings of the IEEE Conference on Computer Vision and Pattern Recognition (CVPR), pp. 1521–1528.
- Vaswani, A., Shazeer, N., Parmar, N., Uszkoreit, J., Jones, L., Gomez, A.N., Kaiser, Ł., Polosukhin, I., 2023. Attention is all you need. In: Advances in Neural Information Processing Systems, Vol. 30, pp. 5998–6008.
- Weinstein, B.G., 2018. A computer vision for animal ecology. *J. Anim. Ecol.* 87, 533–545. https://doi.org/10.1111/1365-2656.12780.
- Wernberg, T., Krumhansl, K., Filbee-Dexter, K., Pedersen, M.F., 2019. Status and trends for the world's kelp forests. In: Sheppard, C. (Ed.), World Seas: An Environmental Evaluation, 2nd ed. Academic Press, London, pp. 57–78. https://doi.org/10.1016/B978-0-12-805052-1.00003-6
- Williams, I.D., Couch, C.S., Beijbom, O., Oliver, T.A., Vargas-Ángel, B., Schumacher, B.D., Brainard, R.E., 2019. Leveraging automated image analysis tools to transform our capacity to assess status and trends of coral reefs. *Frontiers in Marine Science* 6, 222. https://doi.org/10.3389/fmars.2019.00222
- Zuo, X., Jiang, J., Shen, J., and Yang, W., 2025. Improving underwater semantic segmentation with underwater image quality attention and multi-scale aggregation attention. Pattern Analysis and Applications 28, 80.

## Supplementary/Appendix Tables

**Table. A1.** Summary of the study sites used for model development, evaluation, and external validation, including the mean kelp cover (%) calculated from binary segmentation masks and the number of image–mask pairs available for each site.

| S.No | Site name | Sampling Year | Lat | Lon | Min. depth (m) | Max. depth (m) | No of image pairs | Mean kelp cover (%) |
|---|---|---|---|---|---|---|---|---|
| Development Dataset (*-valid, **-test and other- train) | | | | | | | | |
| 1 | Bakers | 2024-06-24 | 42.537 | -70.786 | 3.1 | 10.4 | 32 | 26.3 |
| 2** | Bald Porcupine | 2024-08-08 | 44.387 | -68.181 | 3 | 16.4 | 39 | 43.4 |
| 3 | Great duck | 2024-08-07 | 44.150 | -68.247 | 3.3 | 8.5 | 108 | 36.3 |
| 4* | Great Gull | 2024-07-15 | 41.202 | -72.118 | 3.4 | 6.2 | 247 | 30.9 |
| 5 | North dumpling | 2024-07-15 | 41.287 | -72.018 | 3.5 | 11.9 | 279 | 23.7 |
| 6 | North Gooseberry | 2022-07-08 | 42.526 | -70.793 | 1 | 5 | 1695 | 25.0 |
| 7** | Ram Island | 2024-07-16 | 41.313 | -71.979 | 1.4 | 6.4 | 33 | 7.71 |
| 8 | South Dumpling | 2024-07-15 | 41.252 | -72.015 | 3 | 10.2 | 339 | 5.14 |
| 9 | V7 (Gooseberry) | 2022 | 42.524 | -70.792 | 1 | 5 | 623 | 9.27 |
| | | | | | | | | |
| Evaluation Dataset | | | | | | | | |
| 1 | Bakers Gooseberry | 2022-07-26 | 42.524 | -70.792 | 1 | 5 | 50 | 28.5 |
| 2 | Bakers Island | 2022-07-20 | 42.531 | -70.783 | 1 | 5 | 50 | 25.1 |
| 3 | Bakers Island | 2022-08-04 | 42.531 | -70.790 | 1 | 5 | 50 | 48.3 |
| 4 | Childrens Island | 2022-07-20 | 42.515 | -70.819 | 1 | 5 | 50 | 12.5 |
| 5 | Castle Rock | 2022-07-03 | 42.496 | -70.831 | 1 | 5 | 50 | 40.7 |
| 6 | Dry Breakers | 2022-07-05 | 42.522 | -70.790 | 1 | 5 | 50 | 8.5 |
| 7 | Dry Breakers | 2022-07-21 | 42.526 | -70.787 | 1 | 5 | 50 | 14.6 |
| 8 | Dry Breakers | 2022-07-27 | 42.519 | -70.784 | 1 | 5 | 50 | 1.1 |
| 9 | Dry Breakers | 2022-08-24 | 42.522 | -70.784 | 1 | 5 | 50 | 18.9 |
| 10 | Gooseberry | 2022-07-21 | 42.527 | -70.777 | 1 | 5 | 50 | 21.4 |
| 11 | Great Haste | 2022-07-01 | 42.533 | -70.843 | 1 | 5 | 50 | 32.6 |
| 12 | Juniper Beach | 2022-07-01 | 42.534 | -70.863 | 1 | 5 | 50 | 37.1 |
| 13 | North Gooseberry | 2022-07-20 | 42.524 | -70.794 | 1 | 5 | 50 | 24.4 |
| 14 | North Gooseberry | 2022-07-27 | 42.525 | -70.793 | 1 | 5 | 50 | 4.4 |
| 15 | North Gooseberry | 2022-08-04 | 42.525 | -70.792 | 1 | 5 | 50 | 38.7 |
| 16 | North Gooseberry | 2022-08-24 | 42.524 | -70.793 | 1 | 5 | 50 | 5.2 |
| 17 | Ram Island | 2022-07-18 | 42.542 | -70.792 | 1 | 5 | 50 | 6.4 |
| 18 | South Gooseberry | 2022-07-27 | 42.522 | -70.790 | 1 | 5 | 50 | 5.1 |
| 19 | South Gooseberry | 2022-08-24 | 42.522 | -70.790 | 1 | 5 | 50 | 1.5 |
| | | | | | | | | |
| External Dataset (USA and Canada) | | | | | | | | |
| 1 | Woody Island (USA) | 2022 | 44.449 | -63.713 | 0.5 | 10 | 43 | 36.4 |
| 2 | Charlotte (USA) | 2022 | 45.312 | -60.968 | 0.5 | 10 | 37 | 39.3 |
| 3 | New Hampshire (CA) | 2014 | 42.972 | -70.620 | 1 | 5 | 50 | 16.5 |

**Table. A2.** Common training settings used for all three deep learning models (ResNet34–U-Net, ResNet50–DeepLabV3, and ASPP–Transformer) for kelp segmentation from underwater RGB images.

| Parameter | Value |
|---|---|
| Input patch size | 512 × 512 |
| Training Epochs | 60 |
| Optimizer | AdamW |
| Learning rate scheduler | Cosine annealing |
| Loss function | BCE + Dice (Weight = 0.5) |
| Data augmentation | Flip, rotation, affine transformation, brightness/contrast adjustment, and Gaussian noise |
| Normalization | ImageNet mean and standard deviation |
| Evaluation metrics | Dice coefficient (primary), IoU |
| Threshold selection | Optimized using validation set predictions (0.05–0.95) |
| Random seed | 42 |

**Table. A3.** Quantitative evaluation of three semantic segmentation models for underwater kelp detection on the train/validation dataset.

| Model | Pixel Acc | IoU | Precision | Recall | F1-score | Dice | N (Patches) |
|---|---|---|---|---|---|---|---|
| **Train dataset** | | | | | | | |
| ResNet34-U-Net | 0.9102 | 0.5891 | 0.6935 | 0.7276 | 0.6912 | 0.6912 | 77582 |
| ResNet50-DeepLabV3 | 0.9029 | 0.5732 | 0.6858 | 0.7154 | 0.6779 | 0.6779 | |
| ASPP-Transformer | 0.8659 | 0.4538 | 0.6165 | 0.5717 | 0.5499 | 0.5499 | |
| **Validation dataset** | | | | | | | |
| ResNet34-U-Net | 0.8998 | 0.6804 | 0.7904 | 0.8242 | 0.7927 | 0.7927 | 18335 |
| ResNet50-DeepLabV3 | 0.8951 | 0.6661 | 0.7853 | 0.8105 | 0.7819 | 0.7819 | |
| ASPP-Transformer | 0.8774 | 0.6275 | 0.7655 | 0.7784 | 0.7480 | 0.7480 | |

## Supplementary/Appendix Figures

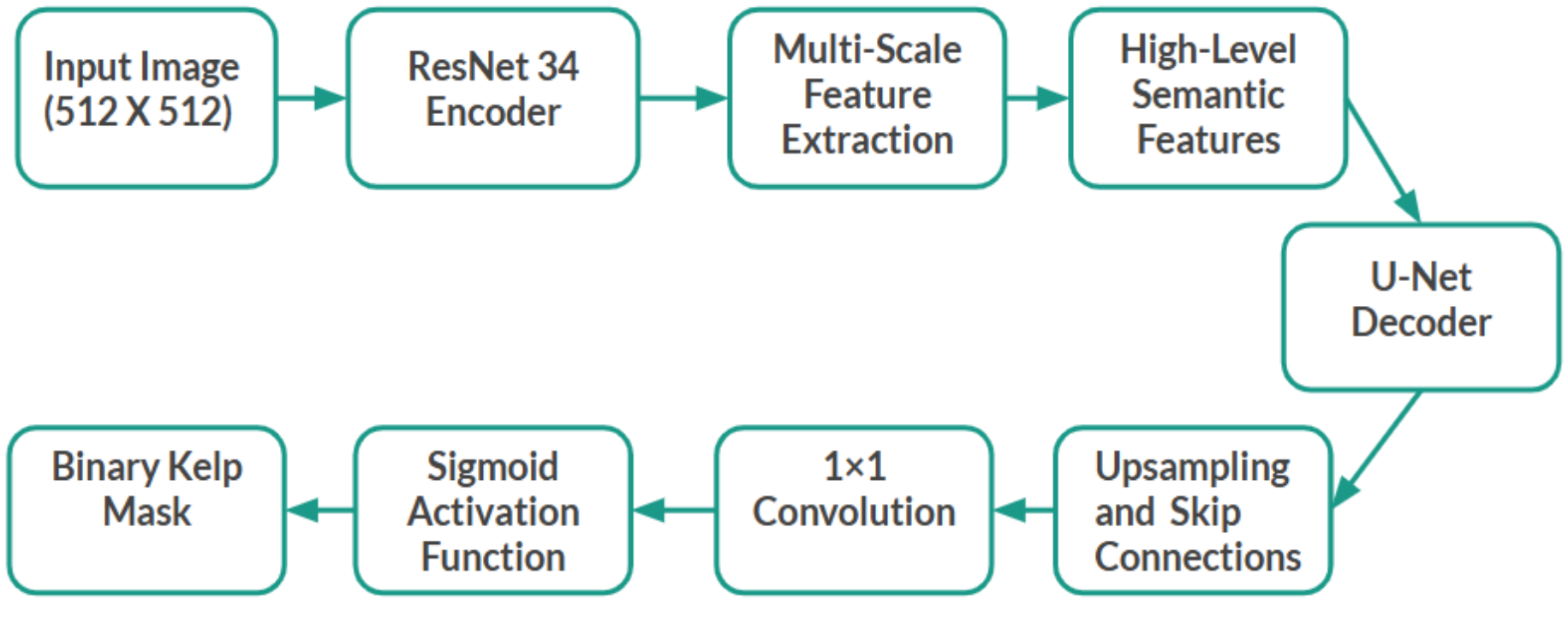

**Figure A1.** Resnet34-U-net architecture

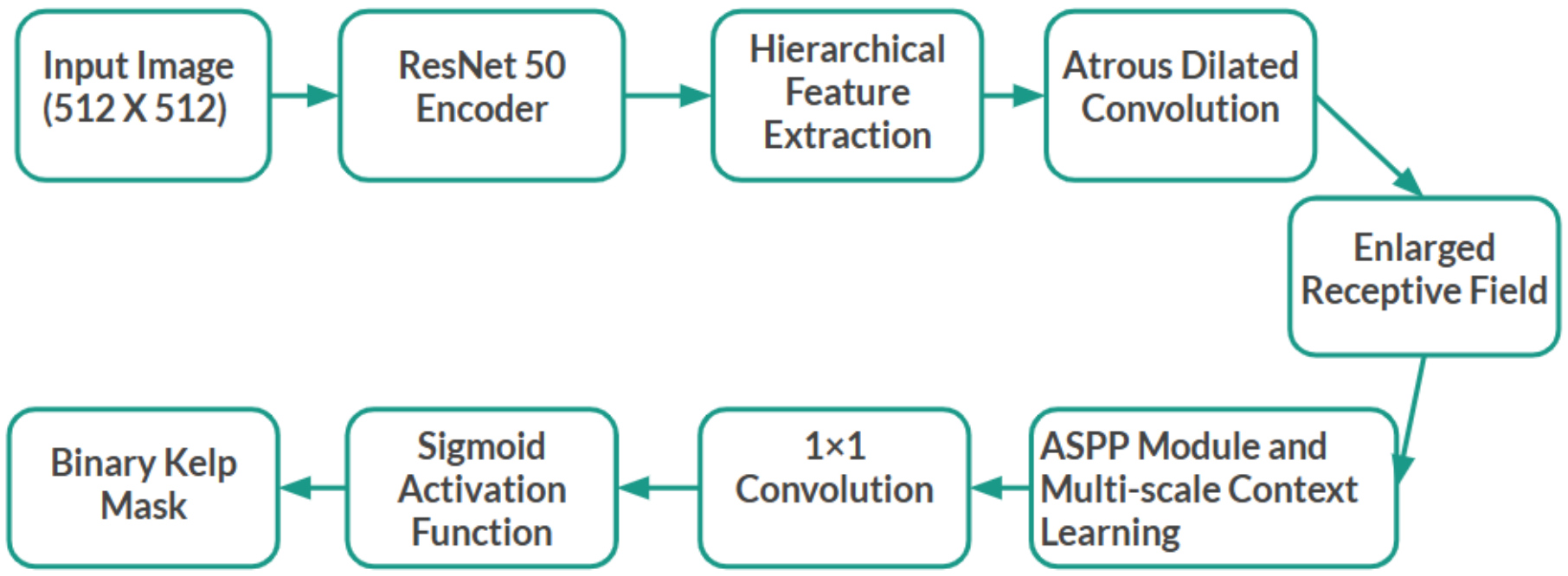

**Figure A2.** Resnet50 -DeeplabV3 architecture

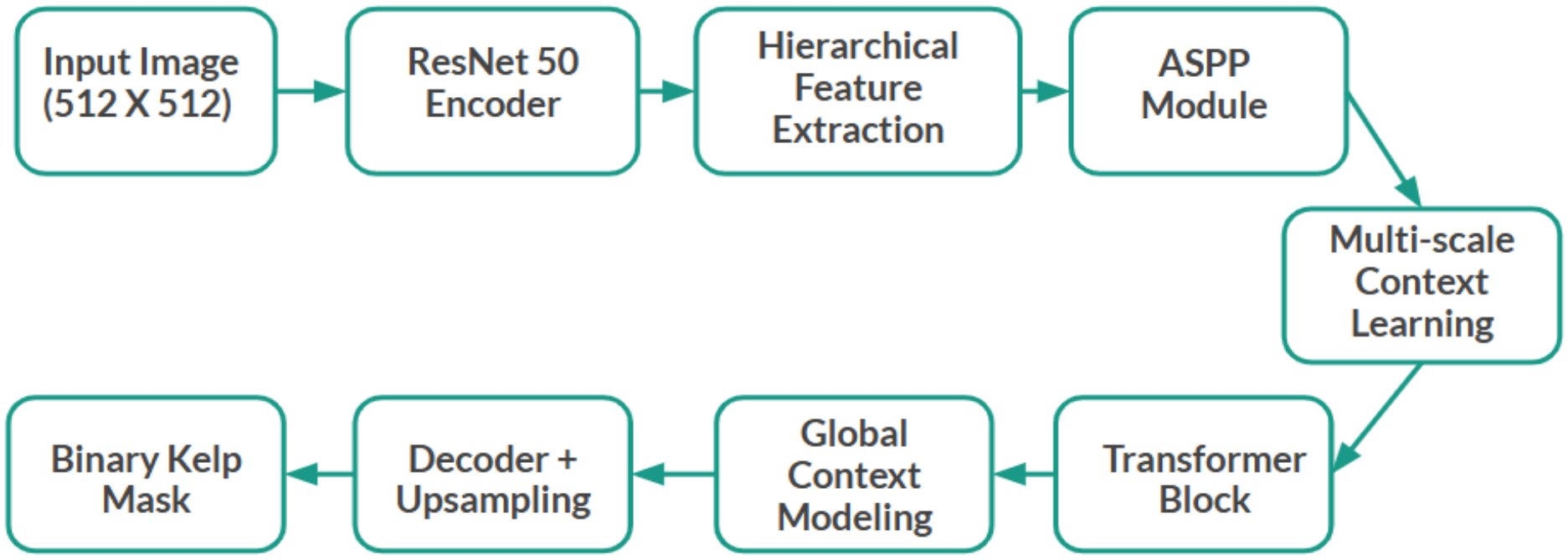

**Figure A3.** ASPP-Transformer architecture

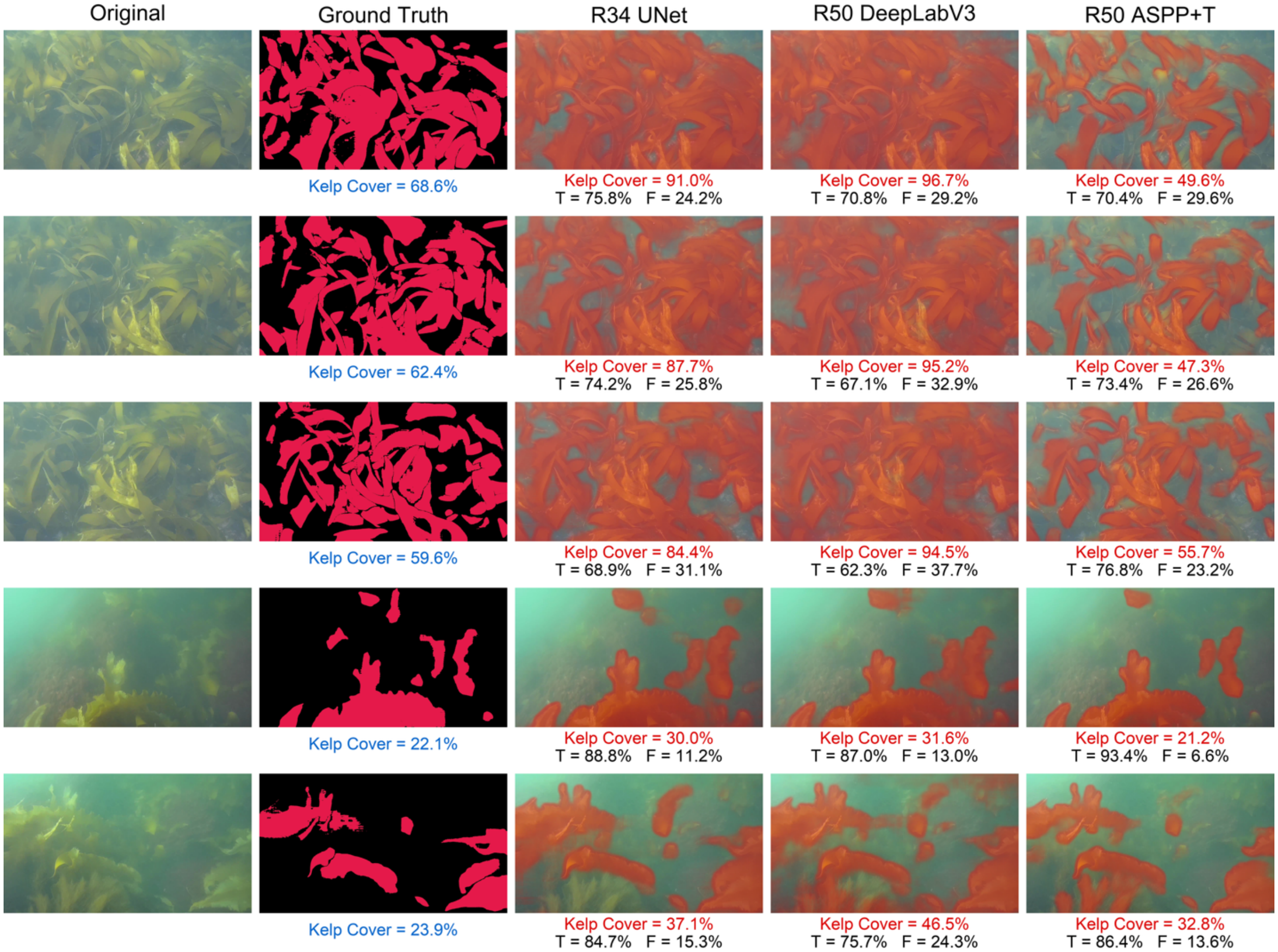

**Figure A4**. Qualitative comparison of kelp segmentation results produced by three deep learning models (ResNet34-U-Net, ResNet50-DeepLabV3, and ASPP-Transformer), shown alongside the original images and ground-truth masks for underwater images used during training and acquired at the Great Duck site. Predicted kelp cover percentages and pixel-wise accuracy statistics are displayed for each mode

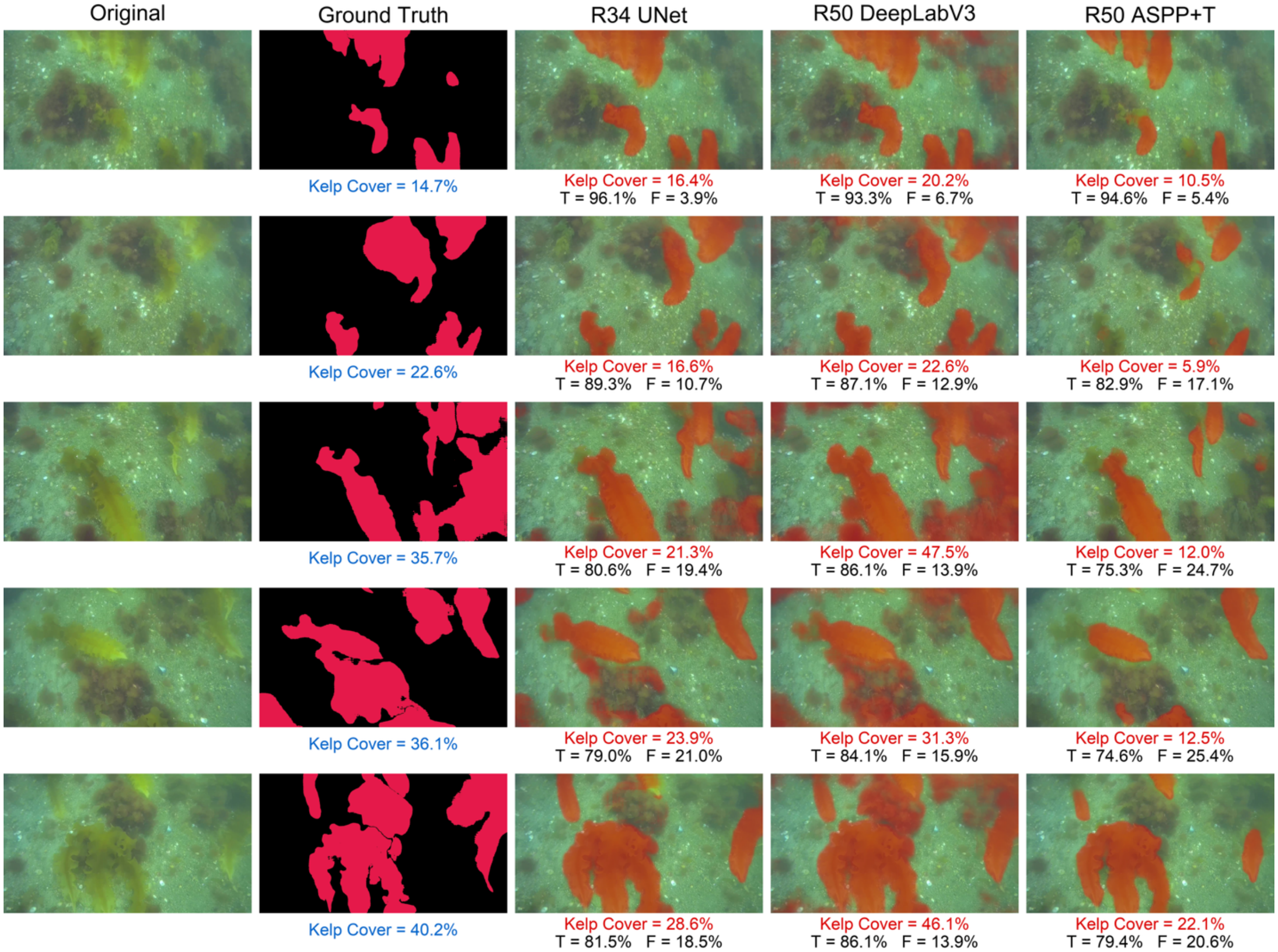

**Figure A5.** Qualitative comparison of kelp segmentation results produced by three deep learning models (ResNet34-U-Net, ResNet50-DeepLabV3, and ASPP-Transformer), shown alongside the original images and ground-truth masks for underwater images used during training and acquired at the Bakers site. Predicted kelp cover percentages and pixel-wise accuracy statistics are displayed for each mode

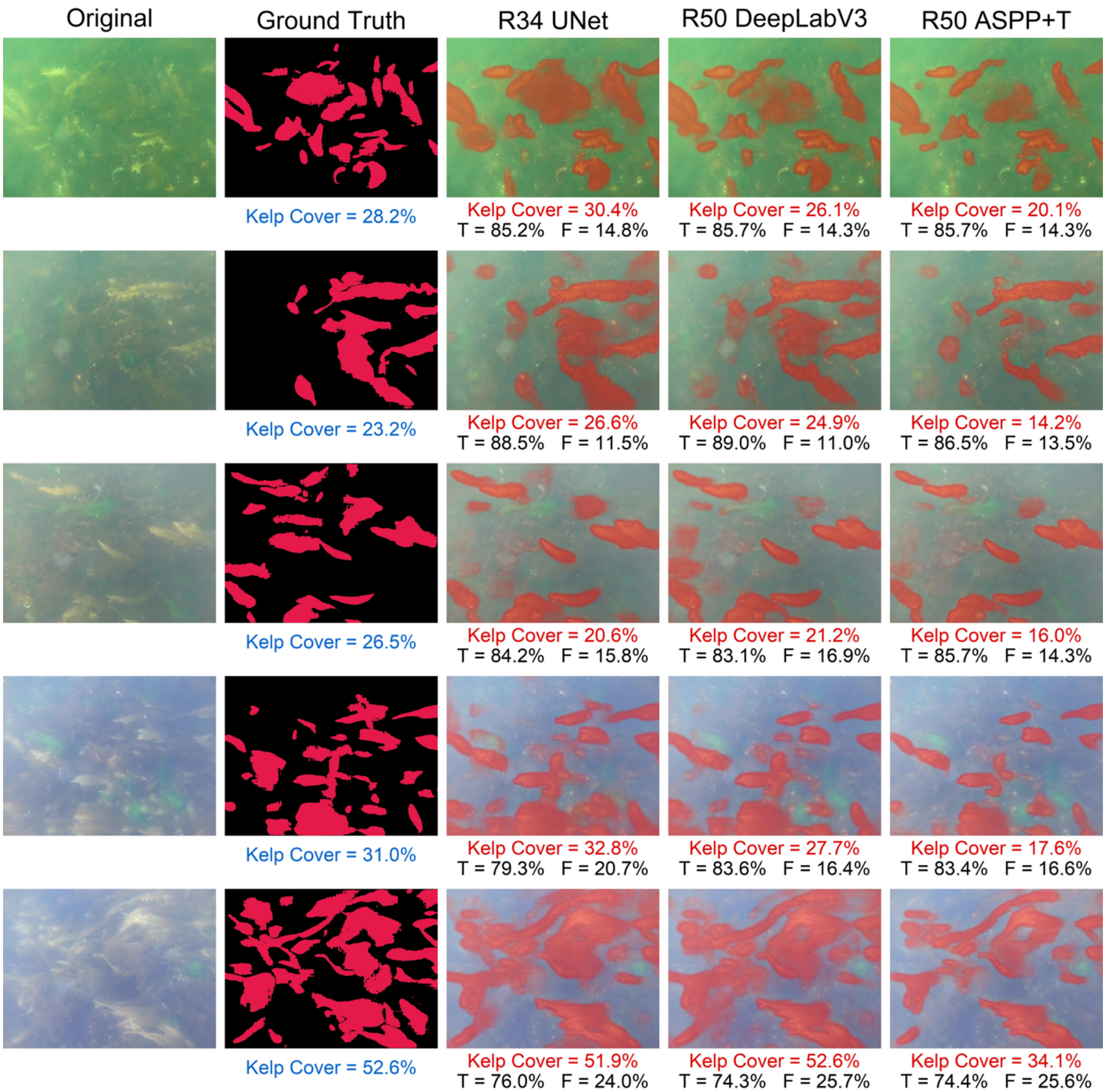

**Figure A6.** Qualitative comparison of kelp segmentation results produced by three deep learning models (ResNet34-U-Net, ResNet50-DeepLabV3, and ASPP-Transformer), shown alongside the original images and ground-truth masks for underwater images used during training and acquired at the North Gooseberry site. Predicted kelp cover percentages and pixel-wise accuracy statistics are displayed for each mode